\documentclass[letterpaper,twocolumn,10pt]{article}
\PassOptionsToPackage{table,xcdraw}{xcolor}
\usepackage{usenix2019_v3}
\usepackage{authblk}

\usepackage{tikz}
\usepackage{amsmath}
\usepackage{dsfont}
\usepackage{enumitem}
\usepackage{amsthm}
\newtheorem{definition}{Definition}
  
\newtheorem{theorem}{Theorem}
\newtheorem{lemma}[theorem]{Lemma}

\usepackage[table,xcdraw]{xcolor}

\usepackage{tablefootnote}
\usepackage{tabularx}
\usepackage{caption}
\usepackage{subcaption}
\usepackage{threeparttable}
\usepackage{booktabs}
\usepackage{multirow}
\usepackage{arydshln}

\usepackage{filecontents}

\begin{document}
%-------------------------------------------------------------------------------

%don't want date printed
\date{}

\title{\Large \bf Towards Lifecycle Unlearning Commitment Management: \\ Measuring Sample-level Unlearning Completeness}

\author{
{\rm Cheng-Long Wang$^{1,2}$, 
Qi Li$^{1,2,3}$, 
Zihang Xiang$^{1,2}$,
Yinzhi Cao$^4$, 
Di Wang\thanks{Corresponding author.}$\:\;^{1,2}$} \\
$^1$Provable Responsible AI and Data Analytics Lab\\
$^2$King Abdullah University of Science and Technology\\
$^3$National University of Singapore\ \ 
$^4$Johns Hopkins University\\
\{chenglong.wang, zihang.xiang, di.wang\}@kaust.edu.sa,
liqi@u.nus.edu,
yinzhi.cao@jhu.edu}

%for single author (just remove % characters)
% \author{
% {\rm Your N.\ Here}\\
% Your Institution
% \and
% {\rm Second Name}\\
% Second Institution
% % copy the following lines to add more authors
% % \and
% % {\rm Name}\\
% %Name Institution
% } % end author

\maketitle

%-------------------------------------------------------------------------------
\begin{abstract}
%-------------------------------------------------------------------------------
% Not more than 200 words, if possible, and preferably closer to 150.
Growing concerns over data privacy and security highlight the importance of machine unlearning—removing specific data influences from trained models without full retraining. Techniques like Membership Inference Attacks (MIAs) are widely used to externally assess successful unlearning. However, existing methods face two key limitations: (1) maximizing MIA effectiveness (e.g., via online attacks) requires prohibitive computational resources, often exceeding retraining costs; (2) MIAs, designed for binary inclusion tests, struggle to capture granular changes in approximate unlearning. To address these challenges, we propose the Interpolated Approximate Measurement (IAM), a framework natively designed for unlearning inference. IAM quantifies sample-level unlearning completeness by interpolating the model's generalization-fitting behavior gap on queried samples. IAM achieves strong performance in binary inclusion tests for exact unlearning and high correlation for approximate unlearning—scalable to LLMs using just one pre-trained shadow model. We theoretically analyze how IAM's scoring mechanism maintains performance efficiently. We then apply IAM to recent approximate unlearning algorithms, revealing general risks of both over-unlearning and under-unlearning, underscoring the need for stronger safeguards in approximate unlearning systems. The code is available at \url{https://github.com/Happy2Git/Unlearning\_Inference\_IAM}.
\end{abstract}

%-------------------------------------------------------------------------------
\section{Introduction}
%-------------------------------------------------------------------------------
Machine Learning as a Service (MLaaS) is rapidly expanding into business, workflows, and personal applications, raising concerns about risks from MLaaS providers handling sensitive, polluted, or copyrighted data~\cite{CarliniIJLTZ23,BrownLMST22,xiao2023theory,xiang2024preserving,xiang2025privacy,xiang2024revisiting}. For example, Google was fined €250 million in France for unauthorized use of publisher content in its AI service~\cite{wsj2024google}. Machine unlearning~\cite{CaoY15}, which removes targeted data's influence from models, offers MLaaS providers a critical data compliance solution by avoiding costly full retraining. Exact machine unlearning~\cite{CaoY15,BourtouleCCJTZL21} modifies training pipelines to reduce data lineage crossing, cutting retraining costs by focusing on relevant submodels or checkpoints. Correctly executed, the unlearned model is equivalent to one retrained without the targeted data. However, popular pre-trained models often prevent the pre-implementation of exact unlearning. This gap spurred approximate machine unlearning algorithms~\cite{IzzoSCZ21,GolatkarARPS21,NeurIPS21Adaptive,MaLLLMR23,cvprMehtaPSR22,abs-2308-07707,abs-2108-11577} that relax unlearning objectives for resource-efficient post-hoc solutions. A model is considered approximately unlearned as long as it is close to the exact retrained model in the parameter space.
 
While approximate unlearning offers fast adaptability, a key question is how much unlearning completeness it achieves. Cao et al.~\cite{CaoY15} first introduced the concept of \textit{unlearning completeness}, which requires that a complete unlearned system gives the same prediction result as the retrained system. They measure this by calculating the percentage of test samples that receive the same prediction results in both the unlearned and retrained models using a representative test set. However, this metric (e.g., test accuracy comparisons) measures unlearning completeness through aggregate model behavior, but cannot isolate the specific influence of individual training samples—the core requirement for precise unlearning verification.
 
We define \textit{sample-level unlearning completeness} as the degree a training sample is unlearned during the unlearning process. Specifically, when a query example is fully unlearned, the unlearned model's response should rely entirely on its generalization ability; otherwise, it returns a fitted output. Sample-level unlearning completeness is the extent a model's behavior for a sample shifts from fitting to generalization. To explain the risks of imprecise unlearning, we introduce the concepts of \textit{under-unlearning}, meaning incomplete removal of unlearned data's impact, and \textit{over-unlearning}, meaning unlearning's detrimental effect on retained data. It is important to note that under-unlearning and over-unlearning are not necessarily mutually exclusive; as they refer to different sets of data (unlearned and retained, respectively), both phenomena can occur to varying degrees simultaneously. Both pose significant risks if not properly measured. The former typically leaves residual private information of unlearned samples in the unlearned model, while the latter may undermine model performance in unpredictable ways, even without external threats. These are distinct but potentially concurrent risks.

Some works reveal threats from the opacity of approximate unlearning~\cite{DiDAKS22, abs-2309-08230, abs-2403-07362,chen2025fedmua}. Recent studies from empirical~\cite{BasuPF21} and theoretical sides~\cite{ChourasiaS23} point out the fragility of these algorithms' relied-upon foundational techniques, specifically in deep learning, even without external threats. To minimize unlearning risks, a good measurement should effectively infer the sample-level unlearning completeness for each sample in the original training set.

The machine unlearning community~\cite{LiuT20,HuangLL21,GravesNG21,MaLLLMR23,GolatkarECCV20} commonly adapts Membership Inference Attacks (MIAs) to evaluate if specific samples are successfully unlearned, as the binary ground truth of exact unlearning aligns with the traditional targets of MIA. However, we argue that MIAs are ill-suited for unlearning inference—the task of quantifying how completely samples are removed—for three reasons: 
i) Offline MIAs~\cite{ShokriSSS17,Ye0001MMBS22} train shadow models solely on population data without the queried sample. These methods tend to misclassify high-confidence non-members (e.g., samples easily generalized by the target model with high confidence) as members, leading to high False Positive Rates (FPR). This high FPR, in turn, reduces the MIA's True Negative Rate (TNR) and can falsely signal under-unlearning (i.e., incorrectly concluding the sample was not unlearned). 
ii) Recent Online MIAs~\cite{CarliniCN0TT22,ZarifzadehLS24} train new shadow models per query using datasets that include the queried sample. Although such methods improve the identification of high-confidence non-members, their computational cost scales linearly with the number of queries, making them impractical for unlearning inference, which is a computationally constrained task. 
iii) Both offline and online MIAs are designed for binary membership tests (in or out). However, such binary ground truth can only appear for exact unlearning. With approximate unlearning, the membership ground truth shifts from a binary status (member vs. non-member) to a more nuanced spectrum, representing varying degrees of data removal or residual influence. These methods then often struggle to capture the granular changes--as our experiments will show--consequently failing to identify under-unlearning or over-unlearning risks.

{\bf Our Work.} For efficient unlearning inference under computational constraints, we propose the Interpolated Approximate Measurement (IAM) method. For each query, IAM interpolate between its responses from the original model (fully fitted) and a few pre-trained shadow OUT models (generalization). This process synthesizes model behavior trajectories from diverse generalization points to a common fitting point, without online training. Trajectory points of different shadow-original model pairs at each interpolation level are treated as samples characterizing the model's response at that fitting level. By applying a Bounded GumbelMap to the signals, the distributions at each level can be effectively fitted using parametric models. This approach also enables parameter estimation with just a single OUT model. For each level's distribution, we compute its Cumulative Distribution Function (CDF) value for the query. This value represents the probability that a response from that distribution is less than the query. The final membership score is derived through weighted averaging of these CDF values. A membership score of 0 means the model's response fully reflects its generalized behavior on the query; as the score approaches 1, the response increasingly reflects the fitted result. Figure~\ref{fig_roc_curve_first} shows the unlearning inference results of IAM on a Receiver Operating Characteristic (ROC) curve. Even with one shadow model, IAM's online and offline variants achieve superior inference performance. Besides superior performance on exactly unlearned samples, IAM is also effective for approximately unlearned samples in various unlearning tasks. The contributions of this paper are summarized as follows:

\begin{figure}[t]
\centering
    \includegraphics[width=\linewidth]{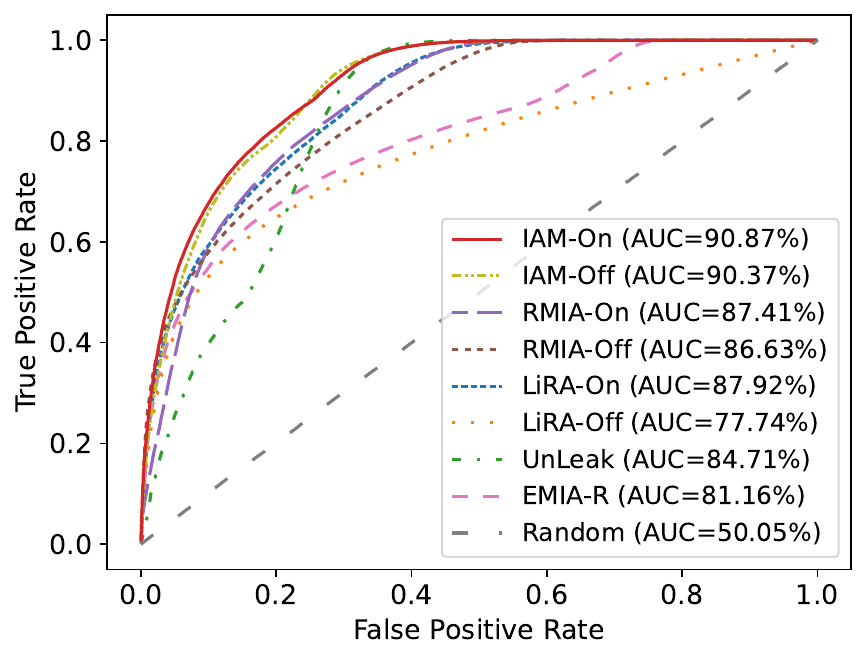}
    \caption{ROC curve of IAM versus prior MIA methods (RMIA~\cite{ZarifzadehLS24}, EMIA~\cite{Ye0001MMBS22}, LiRA~\cite{CarliniCN0TT22}, Unleak~\cite{Chen000HZ21}) for exact unlearning inference on CIFAR-100. Suffixes \textsc{-On} and \textsc{-Off} denote online and offline variants of methods, respectively. Evaluation involved randomly unlearning 500-sample batches (average of 10 runs). All methods are limited to \textbf{one pre-trained shadow model} (used as the OUT model). For online variants, the original model serves as the IN model.}
\label{fig_roc_curve_first}
\end{figure}

\begin{enumerate}
    \item We define the task of unlearning inference and argue existing MIAs are inadequate for this task due to computation and robust quantification issues. We therefore propose Interpolated Approximate Measurement (IAM)—a framework natively designed for this task.
    
    \item Our theoretical analysis explains why IAM's scoring performs efficiently. We establish theoretical bounds on IAM's estimation error using variance analysis and Popoviciu's inequality, confirming IAM can provide reliable parametric estimates for unlearning inference, even with only one shadow model.
    
    \item We benchmark IAM against baselines on exact and approximate unlearning inference tasks. Results show IAM's superior performance, its predicted membership scores strongly correlating with model behavior shifts. We also extend IAM to offline variants and LLM applications, making it applicable for further development. We apply IAM to evaluate approximate unlearning methods, identifying under- and over-unlearning risks across groups and providing insights into their effectiveness.
\end{enumerate}

\section{Background and Related Work}
\subsection{Background}

\noindent {\bf \textit{i) Machine Unlearning. }} Machine unlearning is the process of removing target training data from a machine learning model without fully retraining it. Cao et al.~\cite{CaoY15} first proposed converting learning algorithms into a summation form. In this form, summations consist of transformed data samples using efficiently computable transformation functions, enabling unlearning of sensitive or polluted data by updating the model with these summations. Then many exact unlearning methods~\cite{BourtouleCCJTZL21,UllahM0RA21,DBLP:conf/infocom/SuL23,tao2024communication,wang2023inductive} have been proposed to reduce retraining costs by redesigning the learning process. While these approaches guarantee complete data lineage removal, for pre-trained large models, they often can only be pre-implemented for fine-tuning and are unsuitable for unlearning initial pre-training data of the close-weight model. This limitation has led to the development of approximate machine unlearning solutions~\cite{IzzoSCZ21,GolatkarARPS21,NeurIPS21Adaptive,MaLLLMR23,cvprMehtaPSR22,abs-2308-07707,abs-2108-11577}, which relax the definition of unlearning but can be applied post-training with less computational effort. Consequently, these methods are applicable even to challenging cases like closed-source LLMs~\cite{nipsYaoXL24,NMI25LLMUnlearning}, providing model owners a viable unlearning pathway.

\noindent {\bf \textit{ii) Approximate unlearning.}} Approximate unlearning offers resource-efficient, post-hoc methods to remove data by adjusting model parameters to approximate a state without that data. We categorize these methods into three types based on their model update strategies:

\textit{Log-based retrieval}: 
Amnesiac Unlearning\cite{GravesNG21} subtracts logged parameter updates for batches containing the target data. Thudi et al.\cite{ThudiDCP22} introduce a Standard Deviation Loss to reduce verification error in log-based unlearning. These methods require storing large gradients, making them practical mainly for small, predefined deletions (e.g., temporary access data). However, gradient interdependence makes it difficult to fully remove a sample’s influence on later updates.

\textit{Hessian-based update}: 
Guo et al.\cite{GuoGHM20} propose certified removal for $l_2$ -regularized  linear models via a Newton update approximating leave-k-out loss. Izzo et al.\cite{IzzoSCZ21} use a projective residual update for efficient deletion in linear/logistic models. Fisher Forgetting~\cite{GolatkarCVPR20,GolatkarECCV20} applies noisy Newton updates using the Fisher Information Matrix to minimize KL divergence. While effective in simple models, these methods face challenges with deep networks due to non-convexity, randomness, and scalability.

\textit{Dynamics Masking}: 
Forsaken~\cite{MaLLLMR23} employs a mask gradient generator to iteratively produce mask gradients, prompting neurons to unlearn the memorization of given samples. Selective Synaptic Dampening~\cite{abs-2308-07707} uses Fisher information from training/forgetting data to identify key forget set parameters, then dampens them based on their significance to the forget set versus overall training data. Jia et al.~\cite{JiaLRYLLSL23} explore model sparsification via weight pruning for machine unlearning. 
While these methods efficiently unlearn by masking parameter dynamics of given samples, the overall impact on model parameters is difficult to predict, relying on complex external evaluation to determine the model utility.

\noindent \textbf{Necessity of Unlearning Measurement.} Unlike exact unlearning's definitive results, the link between approximate parameter-space operations and sample-level unlearning completeness is not well understood. For instance, whether these operations are linear, nonlinear, or influenced by unknown dependencies remains unclear~\cite{elhage2021framework}. Consequently, approximate unlearning algorithms often cannot precisely control sample-level unlearning completeness outcomes. A measurement step is thus essential to reduce unlearning risks.

\subsection{Related Work}

\noindent {\bf \textit{i) Dataset Auditing.}} 
Dataset auditing verifies if a query dataset was removed from a model, often in a black-box setting accessing only model outputs, contrasting with MIA's focus on individual samples. Recent methods tackle challenges here; for example, the calibrating approach~\cite{LiuT20} mitigates MIA false positives using a calibrated model (trained on similar, non-overlapping data) and applies the K-S distance to determine if the target model has used or forgotten the query dataset. Ensembled Membership Auditing (EMA)~\cite{HuangLL21} uses a two-stage process: first applying sample-level MIAs with various metrics, then thresholding and aggregating results into a binary dataset decision. However, these methods typically infer the dataset membership status as a whole, overlooking the sample-level auditing needed to mitigate sample-level unlearning anomalies. Such anomalies occur when unlearning effects are inconsistent across data points (e.g., some retained samples are disproportionately impacted by the unlearning process, termed sample-level over-unlearning), issues potentially masked by aggregate evaluations.

\noindent {\bf \textit{ii) Proof-of-(Un)Learning.}} Proof-of-learning (PoL) is a mechanism allowing a model creator to generate proof of the computational effort required for training. Jia et al.~\cite{JiaYCDTCP21} first proposed a PoL solution logging training checkpoints to ensure spoofing is as costly as obtaining the proof via actual model training.
However, Thudi et al.~\cite{thudi2022necessity} point out that the PoL framework is unsuitable for auditing unlearning, as one can spoof approximate unlearning verification without any model modification. Recent work~\cite{WengYDHWW24, ChoiSD23} thus addresses spoofing threats in PoL and Proof-of-Unlearning (PoUL) using verifiable unlearning protocols adaptable to various algorithms. However, Po(U)L, focusing on proving computational execution in the (Un)-Learning process (whether exact or approximate), differs from unlearning completeness measurements addressing honest unlearning. Work on PoL/PoUL overlooks that for many approximate unlearning algorithms, even without external threats, faithful execution doesn't guarantee the complete removal of data lineage due to relaxed unlearning criteria. Discussions related to attacks on Proof-of-(Un)Learning, such as forging attacks, fall outside our scope.

\noindent {\bf \textit{iii) Data Memorization.}}
Data Memorization describes machine learning models retain and potentially reveal specific details from their training data, rather than merely capturing general patterns. 
Arpit et al.~\cite{DBLP:conf/icml/ArpitJBKBKMFCBL17} show DNNs' memorization degree depends not only on the architecture and training procedure but also on the training data itself. Feldman~\cite{DBLP:conf/stoc/Feldman20} formalizes and quantifies sample-level \textit{leave-one-out} memorization, analyzes its necessity for achieving optimal generalization. For each labeled training sample, its memorization score is determined by the expectation of the prediction accuracy drop when excluded from training.

\section{Revisiting MIAs for Unlearning Inference}\label{sec:subsec:measurement}

This section introduces important definitions for unlearning inference tasks and unlearning risks; a notation table is in Appendix~\ref{appendixnotation}. Given MIA aligns with the ground truth of exact unlearning, and the community~\cite{LiuT20,HuangLL21,GravesNG21,MaLLLMR23,GolatkarECCV20} often repurposes MIAs to evaluate unlearning success, this section begins by revisiting MIA. Subsequently, we formalize unlearning inference tasks for exact and approximate scenarios, define under- and over-unlearning, and detail the gap when adapting MIAs for this purpose.

\begin{definition}[\textbf{Membership Inference Attack}]
Let $\mathcal{D}$ and $\mathcal{A}$ be the data distribution and learning algorithm. The game proceeds as follows: 
\begin{enumerate}[leftmargin=0.8cm]
    \item \textbf{Learning Phase:} The challenger samples a training dataset $D \sim \mathcal{D}$, and train a model $\theta \sim \mathcal{A}(D)$ on $D$.
    
    \item \textbf{Challenge Creation}: The challenger flips a bit $b$. If $b = 1$, it selects $z$ from the training dataset $D$. If $b = 0$, it samples $z$ from the population $\mathcal{D}$ such that $z \notin D$. The challenger then sends $z$ and $\theta$ to the adversary.
    
    \item \textbf{Adversary's Task}: The adversary, having access to the data distribution $\mathcal{D}$ and model $\theta$, infers the bit $\hat{b}$.

\end{enumerate}

\end{definition} 

Score-based MIAs~\cite{ZarifzadehLS24,Ye0001MMBS22,CarliniCN0TT22} first compute a prediction $\text{Score}(z; \theta)$ and the binary decision on $\hat{b}$ is determined by:
\begin{equation}
    \hat{b} = \mathds{1}[\text{Score}(z; \theta) > \tau],
\end{equation}
where $\tau$ is a threshold that controls the trade-off of FPR and False Negative Rate (FNR). In a classification task, the above game assumes the adversary has access to both the query example and its ground-truth label $z=(x,y)$.

To estimate $\text{Score}(z; \theta)$ and determine a suitable threshold $\tau$, most MIAs train shadow models on population data to simulate the generalization and fitting behavior of the query example, then derive an effective score function. More shadow models enable more accurate capture of the model behavior pattern. The performance of MIAs can be evaluated using the Area Under the ROC Curve (AUC-ROC), which measures how well predicted outputs rank members above non-members. Carlini et al.~\cite{CarliniCN0TT22} emphasize the importance of reliably identifying worst-case privacy leakage. Thus, MIA evaluation often focuses on a specific region of the ROC curve: the True Positive Rate (TPR) at low FPR.

\begin{definition}[\textbf{Binary Unlearning Inference}] 
Denote $\mathcal{A'}$ as an exact unlearning algorithm for $\mathcal{D}$ and $\mathcal{A}$.

\begin{enumerate}[leftmargin=0.8cm]
    \item \textbf{Learning Phase:} A challenger samples a training dataset $D \sim \mathcal{D}^n$ and trains an original model $\theta \leftarrow \mathcal{A}(D)$.
    \item \textbf{Unlearning Request:} The challenger receives an unlearning bit vector $\mathbf{b} \in \{0,1\}^n$, where $\mathbf{b}_i=0$ indicates $z_i \in D$ is targeted for unlearning. Typically, unlearning requests target small batches of data.
    \item \textbf{Exact Unlearning Phase:} The challenger applies $\mathcal{A'}$ to $\theta$ producing an exact unlearned model $\theta' \leftarrow \mathcal{A'}(\theta, D, \mathbf{b})$. The challenger then sends $(D, \theta, \theta')$ to the verifier.
    \item \textbf{Verifier's Task:} The verifier, using $D, \theta, \theta'$ (and potentially knowledge of $\mathcal{D}$), infers the unlearned bit $\hat{\mathbf{b}}$, where $\hat{\mathbf{b}}_i$ denotes the predicted unlearning result of $z_i \in D$.
 
\end{enumerate}
\label{def:exact}
\end{definition} 

Only when exact unlearning is applied, per-sample unlearning inference becomes a binary inference task: determining if the sample was \textit{exactly} removed from the training set. In such cases, binary decisions from MIA directly transfer to \textbf{Bin}ary \textbf{U}nlearning \textbf{I}nference (\textbf{BinUI}). The distinction lies in their computational requirements and evaluation focus.

Powerful MIAs often train numerous shadow models; costly online attacks may even train new IN models per query to detect worst-case privacy leakage. For unlearning inference, as unlearning aims to avoid retraining costs, its measurement must also avoid costs exceeding retraining. Fortunately, the original model $\theta$ can serve as a no-cost, pre-trained IN model, capturing its fitting behavior for training samples. Following MIA's naming convention, we term methods using the pre-trained IN model ($\theta$) and pre-trained OUT models as \textbf{online inference}, and those relying only on pre-trained OUT models as \textbf{offline inference}. Since $\theta$ is the sole pre-trained IN model used (no new ones are trained), both inference approaches are practical with limited pre-trained OUT models.

Unlearning inference aims to minimize under- and over-unlearning risks. Unlike MIA's focus on extreme low-FPR regions (distinguishing highly confident members), BinUI evaluation requires precise unlearning completeness alignment for all $z \in D$ to correctly rank both retained (members) and unlearned (non-members) examples. Full AUC-ROC is thus ideal for BinUI evaluation, holistically assessing TPR (for retained members) and TNR (for unlearned samples) across all scores. Conversely, MIA's TPR, used for worst-case privacy leakage detection, is ill-suited for identifying poorly unlearned examples. Such a high TPR might merely reflect numerous correctly identified retained samples, not necessarily capturing the worst-case unlearned example, which could retain some member-like confidence.

We now extend unlearning measurement to the approximate unlearning setting. Unlike exact unlearning, approximate methods typically suppress outputs for unlearned samples via global or local parameter updates, aiming to be close to the ground-truth unlearned model's parameter space. However, the link between parameter updates and individual sample influence isn't fully understood; for example, recent work~\cite{elhage2021framework} shows a given internal model representation can map to multiple inputs. Consequently, a local parameter update could affect model behavior for both retained and unlearned samples, possibly to different degrees. To denote the non-binary ground truth of approximate unlearning, we define the \textit{unlearning completeness} score and the phenomena of \textit{under-} and \textit{over-unlearning}.

\begin{definition}[\textbf{Sample-level Unlearning Completeness and Membership Score}]  
For a data point $ z_i $ and target model $\theta$, we define the membership score $ s_i \in [0, 1] $ and unlearning completeness $ (1 - s_i) $ as a pair of complementary metrics quantifying the closeness of the model’s behavior on $ z_i $ to the two ends of the generalization-fitting spectrum. Specifically: $s_i = 1$ indicates $z_i$ is fully retained (model outputs remain completely fitted to $z_i$); $s_i = 0$ signifies $z_i$ is completely unlearned (model responses rely purely on generalization).\footnote{Since these terms \textit{sample-level unlearning completeness} and \textit{membership score} are complementary, we will use them interchangeably in the following of the paper.}   

\end{definition}  

Notably, the opacity of approximate algorithms makes it challenging to directly ascertain the unlearning completeness $\mathbf{s}$ for all $z\in D$, even for model owners and algorithm executors. We therefore rely on measurement techniques to produce a predicted unlearning score $\hat{\mathbf{s}}$.

\begin{definition}[\textbf{Score-based Unlearning Inference Game}] \ 
Denote $\tilde{\mathcal{A}}$ an unlearning algorithm (exact or approximate) for $\mathcal{D}$ and $\mathcal{A}$.

\begin{enumerate}[leftmargin=0.8cm]
    \item {The first two steps are the same as steps 1) and 2) in Definition~\ref{def:exact}}. 
    
    \item \textbf{Approximate Unlearning Phase}: The challenger applies $\mathcal{A'}$ to produce an approximately unlearned model $\theta' \sim \tilde{\mathcal{A}}(D, \mathbf{b}, \mathcal{A}, \theta)$. The challenger sends ($D$, $\theta$, $\theta'$) to the verifier.
    
    \item \textbf{Verifier's Task}: The verifier, using $D, \theta, \theta'$ (and potentially knowledge of $\mathcal{D}$), infers the score vector $\hat{\mathbf{s}}$, where $\hat{\mathbf{s}}_i\in [0,1]$ is the predicted membership score for $z_i\in D$. 
    
\end{enumerate}    
\end{definition}

Intuitively, membership scores from score-based MIAs can be applied to \textbf{Score}-based \textbf{U}nlearning \textbf{I}nference (\textbf{ScoreUI}). Specifically, a high MIA membership score corresponds to low unlearning completeness, and vice versa. Similar to MIA's application to BinUI, adapting score-based MIA for ScoreUI also requires reconsidering the computational requirements and evaluation focus.

\begin{figure*}[t]
\centering
    \includegraphics[width=\linewidth]{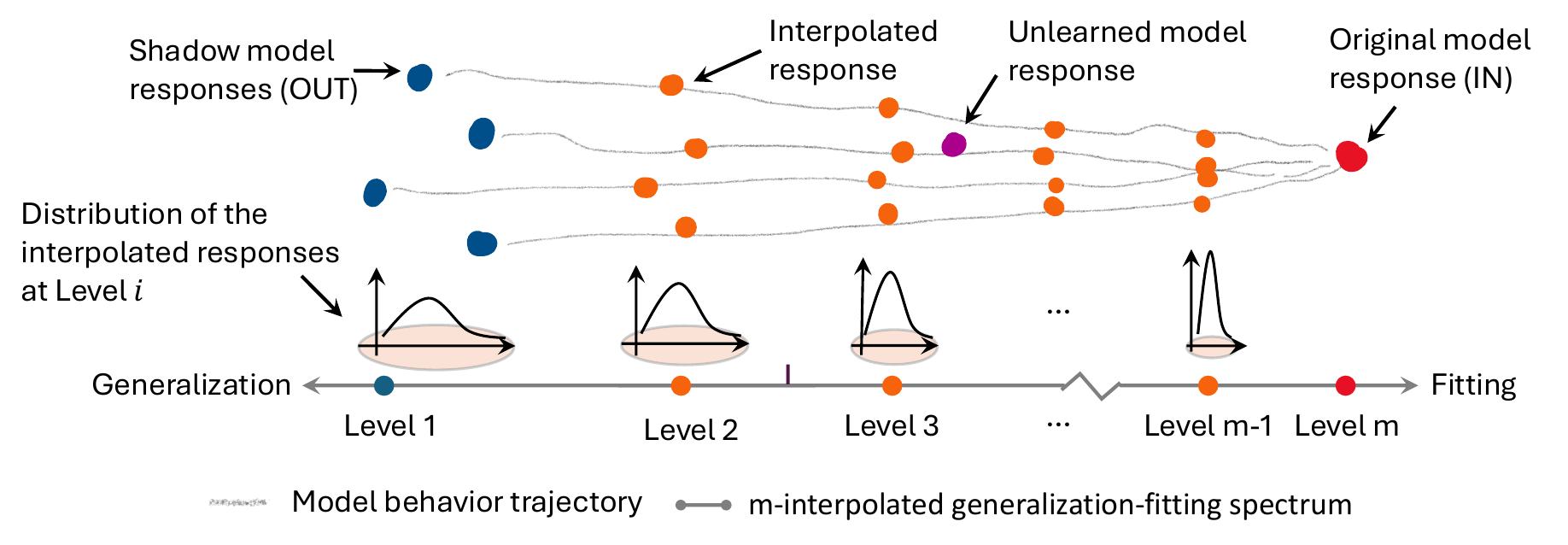}
    \caption{IAM Framework. Observe the figure top-down to understand its steps. The top portion shows the interpolation sketch: blue \textit{Shadow model responses} (representing OUT model generalization behaviors) and the red \textit{Original model response} (the fitted IN model state) are connected by grey \textit{Model behavior trajectories}. For each trajectory (representing a path on the generalization-to-fitting spectrum starting from a different generalization state), linear interpolation along it creates orange \textit{Interpolated responses}. Next, IAM estimates the distribution of these responses (collected from all trajectories) at each interpolated level using parametric models and finally aggregates results from all levels for a final score.}
\label{fig_framework}
\end{figure*}

The evaluation of an approximate unlearning has two aspects: the unlearning gap $\Delta_{\mathbf{b}, \mathbf{s}}$ between $\mathbf{b}$ and $\mathbf{s}$, and the inference gap $\Delta_{\mathbf{s}, \hat{\mathbf{s}}}$ between $\mathbf{s}$ and $\hat{\mathbf{s}}$. A smaller $\Delta_{\mathbf{b}, \mathbf{s}}$ indicates better unlearning performance, and a smaller $\Delta_{\mathbf{s}, \hat{\mathbf{s}}}$ reflects more accurate inference performance. Effective approximate unlearning should keep the membership score of retained samples near $1$ and the score of unlearned samples near $0$. 

The gap $\Delta_{\mathbf{b}, \mathbf{s}}$ is quantified using Binary Cross-Entropy (BCE) loss, directly measuring differences between predicted membership scores and target values. Ranking-based metrics like AUC can be misleading here, since they only consider relative score ordering, not actual unlearning completeness. As unlearning requests often involve small batches, we address potential class imbalance between retained and unlearned samples using a weighted BCE loss: 
\begin{equation}
\text{BCE} = -\frac{1}{n} \sum_{i=1}^n \left[w_i\mathbf{b}_i \log(\mathbf{s}_i) + (1 - \mathbf{b}_i) \log(1 - \mathbf{s}_i)\right],
\end{equation}
where $w_i$ is the weight factor that accounts for class imbalance. However, as we have discussed, $\mathbf{s}$ is unavailable although it exists objectively. 
In such cases, we turn to the proxy strategies. For unlearning performance, when the scales of $\mathbf{s}$ and $\hat{\mathbf{s}}$ are aligned and $\Delta_{\mathbf{s}, \hat{\mathbf{s}}}$ is small, $\Delta_{\mathbf{b}, \hat{\mathbf{s}}}$ closely approximates $\Delta_{\mathbf{b}, \mathbf{s}}$, making it a reliable proxy. 

For evaluating the result of ScoreUI, we generate the ground truth $\mathbf{s}$ and then evaluate the performance based on $\Delta_{\mathbf{s}, \hat{\mathbf{s}}}$.~\footnote{Details on the generation of $\mathbf{s} $ are provided in Section~\ref{exp_ScoreUI}.} For $\Delta_{\mathbf{s}, \hat{\mathbf{s}}}$, different measurements may output $\hat{\mathbf{s}}$ in different scales than $\mathbf{s}$. For example, Carlini et al.~\cite{CarliniCN0TT22} uses a likelihood ratio as the predicted score.
Therefore, scale-dependent metrics like MSE loss or L1 distance are not appropriate.
As both $\mathbf{s}, \hat{\mathbf{s}}$ are real-valued vectors, we use Spearman correlation to measure how well the prediction $\hat{\mathbf{s}}$ preserves the ordering of the values. A good metric should assign a higher predicted value to a query with a higher ground truth unlearning completeness, maintaining monotonicity in the relationship.

We then apply the outputs of ScoreUI to identify the sample-level under- and over-unlearning risks, and introduce thresholds for practical identification:

\begin{definition}[\textbf{Under-unlearning}]
\label{def:under}
An unlearned sample $ z_i $ ($ b_i = 0 $) undergoes under-unlearning  if its ground truth score $ s_i > 0 $, indicating $z_i$'s residual influence on the model $\theta$.
\end{definition}

\begin{definition}[\textbf{Over-unlearning}]
\label{def:over}
A retained sample $ z_j $ ($ b_j = 1 $) suffers over-unlearning if its ground truth score $ s_j < 1 $, implying unintended suppression of model $\theta$'s behavior on $z_j$.
\end{definition}

\begin{definition}[\textbf{Threshold-Based Risk Identification}]  
\label{def:thresholds}  
For practical measurement:  
\begin{itemize}[noitemsep,topsep=0pt]  
  \item A sample $z_i$ is identified as posing an under-unlearning risk if its predicted score $\hat{s}_i > \delta_1$, where $\delta_1 \in [0,1]$ is a threshold near 0.  
  \item A retained sample $z_j$ is identified as posing an over-unlearning risk if $\hat{s}_j < \delta_2$, where $\delta_2 \in [0,1]$ is a threshold near 1.  
\end{itemize}  
\end{definition}

\section{Interpolated Approximate Measurement}

To enable reliable measurement for the unlearning inference tasks, we propose the Interpolated Approximation Measurement (IAM) framework, with its core concept illustrated in Figure~\ref{fig_framework}. Given access to the original (IN) model and a limited number of pre-trained shadow (OUT) models,\footnote{The training details of shadow models are provided in Appendix~\ref{shadow_model}.} IAM simulates and captures the model's transition from the generalization behavior exhibited by the OUT models to the fitted state of the IN model. This simulation effectively establishes a generalization-to-fitting spectrum for each OUT-IN model pair. The target model's degree of fitting on the query example is then determined by its position along this spectrum. Such positioning, in turn, helps determine the target model's unlearning completeness for the example.

\subsection{Interpolation Framework}

Here we outline the general framework, with specific functional details presented in the next subsection.

\textbf{Phase 1: Response Interpolation.} 
Analogous to how leading MIA techniques~\cite{CarliniCN0TT22,ZarifzadehLS24} leverage IN/OUT model responses to infer binary membership of examples, a similar understanding of model behavior across diverse learning degrees also helps measure unlearning completeness. The initial phase of IAM aims to simulate model responses along the generalization-to-fitting spectrum, without costly online retraining. As shown in Figure~\ref{fig_framework} (top portion), this simulation is achieved by interpolating responses of the original and pre-trained shadow OUT models. Let $\theta$ be a classification model trained on $D$, and $r(z; \theta)$ its response to a query example $z\in D$.\footnote{Response function will be discussed in Section~\ref{sec:init_response} and~\ref{sec:stable_response}.} Let $\theta'$ denote the unlearned model obtained after applying an unlearning algorithm to $\theta$. $\Theta$ denotes the set of pre-trained shadow OUT models. For any shadow model $\tilde{\theta}\in \Theta$, its response to $z$ is $r(z; \tilde{\theta})$. This response solely reflects the model's generalization behavior. For each shadow-original model pair $(\tilde{\theta}, \theta)$, IAM interpolate between $r(z; \tilde{\theta})$ and $r(z; \theta)$. Let $m \geq 2$ be the number of interpolation steps. The $i$-th interpolated response $r_i(z;\tilde{\theta}, \theta)$ (abbreviated $r_i$) for $i \in \{1, \dots, m\}$ is defined as:
\begin{equation}
\begin{split}
    r_i = \frac{m-i}{m-1}\cdot r(z; \tilde{\theta}) + \frac{i-1}{m-1} \cdot r(z; \theta).
\end{split}
\label{interpolation_equ}
\end{equation}
Clearly, $r_1 = r(z; \tilde{\theta})$ for $i=1$, and $r_m = r(z; \theta)$ for $i=m$. Moreover, the $m$-th responses ($r_m$) from all shadow-original pairs are identical and do not provide a meaningful distribution. Therefore, $r_m$ is not used for subsequent estimation.

\textbf{Phase 2: Statistical Modeling.}

A model’s response to a query typically changes in a predictable and largely monotonic manner as it transitions from generalization to fitting. This is theoretically supported by~\cite[Sec. 8.2.3; Fig. 8.2]{DBLP:books/daglib/0040158}, which describes convergence toward stable, low-cost regions, and empirically by~\cite[Sec. 7.8; Fig. 7.3]{DBLP:books/daglib/0040158}, where training loss steadily decreases during overfitting. For each interpolation step, responses across shadow-original model pairs characterize model behavior at that fitting level. We refer to this collection of responses at step $i$ as Level $i$ responses. Assuming Level $i$ responses follow distribution ${\Phi}_i$, we estimate $q_i$ as the probability that the unlearned model's response $r' = r(z; \theta')$ is larger than the responses from ${\Phi}_i$:
\begin{equation}
q_i = \Pr[r' > X],\ \text{where } X\sim {\Phi}_i.
\label{pdf_calc}
\end{equation}
If the unlearned model's response $r'$ significantly exceeds typical responses from Level $i$ towards the 'fully fitted' end of the spectrum, the corresponding membership score increases, and unlearning completeness decreases. Section~\ref{sec:init_response} details the estimation of ${\Phi}_i$ and calculation of $q_i$.

\textbf{Phase 3: Scoring.} 
 After Phase 2, we obtain $q_i$ values for $i=1,\ldots,m-1$. Simple averaging might seem intuitive, but can result in misleading membership score predictions. For example, if a response is randomly selected from $\Phi_1$ (where its responses reflect only pure generalization behavior), there is approximately a $50\%$ chance this selection could lead to a high $q_1$. Such a high $q_1$ would, under simple averaging, push the estimation towards the fitted end and therefore unreasonably increase the membership score. Therefore, we use weighted averaging, assigning lower weights to earlier levels to mitigate potentially misleading signals:

\begin{equation} 
\text{Score}(z; \theta,\theta') = \frac{\sum_{i=1}^{m-1} i\cdot q_i}{\sum_{i=1}^{m-1} i}.
\end{equation}

When $m=2$, $\text{Score}(z; \theta,\theta') =  \Pr[r' > r _1],\  r_1\sim {\Phi}_1$, where ${\Phi}_1$ represents the response distribution of shadow OUT models. If we replace $\Phi_i$ with a Gaussian distribution, this aligns with the inference stage of offline LiRA\cite{CarliniCN0TT22}. When $m$ increases, the framework captures more intermediate states along the model behavior trajectory. We’ll evaluate the impact of $m$ on membership score estimation in our experiments. 

\subsection{Modeling Interpolated Responses}\label{sec:init_response}

We now discuss designing a good response function whose distribution can be well-approximated by parametric models. Common MIA response functions~\cite{Ye0001MMBS22,ZarifzadehLS24} for a query $z=(x,y)$ are model confidence $p=\Pr[Y=y|X=x;\theta]$ (or $p(z;\theta)$) and cross-entropy loss $\ell(z;\theta) = -\log(p)$. When analyzing the model behaviors for the query example, their confidence distributions for different query samples can exhibit significant variability. For instance, models may yield low confidence for non-member samples less represented by the training data, but higher confidence for those more typical or aligning well with learned general features.~\footnote{Comprehensive statistics of OUT/IN model confidences for all training examples are provided in the Appendix~\ref{out_stat_conf}.} For these high-confidence non-member examples, distinguishing them from actual members can be difficult if their IN-model and OUT-model confidence scores are very similar. Consequently, the scale of the confidence gap between IN and OUT models often varies substantially across examples.

To enable uniform-scale estimation, we apply a double negative logarithm to the cross-entropy loss. Thus, a model $\theta$'s response $r(z;\theta)$ to a query $z$ is calculated as:
\begin{equation} 
r(z;\theta) 
= -\log\Bigl(- \log\bigl(p\bigr)\Bigr),
\label{eq_origin_gumbel}
\end{equation}
where $p$ is the Softmax model confidence. The larger the confidence $p$, the larger the response $r(z;\theta)$.

We refer to the transformation in Eq.~(\ref{eq_origin_gumbel}) as the \textit{GumbelMap}, primarily because its double-log form is inspired by the functional form of the Gumbel distribution~\cite{gumbel1954statistical}. This transformation shows heightened sensitivity as model confidence $p \rightarrow 1$, allowing for more effective differentiation of high-confidence samples. Transformed responses $r(z;\theta)$ can be efficiently approximated using the Gumbel distribution as a parametric model. Let $\mu, \sigma^2$ be the mean and variance of OUT model responses. The Gumbel parameters $\alpha, \beta$, probability density function (PDF), and cumulative distribution function (CDF) are estimated via the method of moments:

\begin{equation} 
\begin{split}
    \alpha & = \mu - \gamma\cdot \beta, \\
    \beta & = \sqrt{\frac{6\sigma^2}{\pi^2}}, \\
    f(x;\alpha,\beta) & = \frac{1}{\beta}e^{-\frac{x-\alpha}{\beta}}e^{-e^{-\frac{x-\alpha}{\beta}}},\\
    F(x;\alpha,\beta) & = e^{-e^{-\frac{x-\alpha}{\beta}}}.
\end{split}
\label{gumbel_cdf}
\end{equation}
where $\gamma\approx 0.5772$ is the Euler–Mascheroni constant. Figure~\ref{fig_hist_signal} (right column) shows estimated Gumbel PDFs for Level 1 responses effectively approximating OUT model responses for all three examples. The method of moments is also computationally efficient, requiring only these means and variances. From all interpolated responses (from shadow-original model pairs), we then compute their mean and variance at each interpolation level. Finally, $q_i$ values (Eq.~\ref{pdf_calc}) are calculated in parallel for all levels.
 
\subsection{Stable Estimation in Extreme Cases}\label{sec:stable_response}
 
In this subsection, we discuss boosting the scoring function in two extreme cases: 1) when a query example is very easy to generalize, its OUT model confidences largely matching the IN model's; and 2) when shadow training resources are limited, yielding few available shadow models.

\noindent {\bf \text{High-confidence Examples.}} A key challenge in membership prediction arises with high-confidence examples, where IN and OUT models exhibit similar behavior: confidences concentrate near 1 and losses approach 0. This occurs because their feature patterns are often typical or already well-captured by the training data. Consequently, this similarity in model confidence makes distinguishing non-members from members difficult. The GumbelMap transformation addresses this through its double-logarithmic mapping property. Specifically, for near-zero loss values, it significantly expands the range towards infinity, amplifying small differences in loss. The closer a loss value is to zero, the greater its amplification. While this transformation enhances member/non-member separation, it has a potential drawback: member queries usually have losses closer to zero than non-members. Consequently, small differences between member queries and the IN model response are amplified more dramatically than their gap with OUT model responses. In other words, it becomes overly sensitive in the near-zero region. 

To mitigate this effect and achieve more robust, meaningful scores for highly fitted examples, we introduce the \textit{Bounded GumbelMap} to update the response function as:

\begin{equation} 
\tilde{r}(z;\theta) 
= -\log\Bigl(\epsilon_1 - \log\bigl(p + \epsilon_2\bigr)\Bigr),
\label{eq_bounded_gumbel}
\end{equation}
where $\epsilon_1,\epsilon_2>0$, and $e^{\epsilon_1}>1+\epsilon_2$. Compared to the original GumbelMap, the \textit{Bounded GumbelMap} has the following properties:
 
\begin{lemma}
For any input $z$, and any model $\theta$, the response $\tilde{r}(z;\theta)$ is bounded. Let $\Theta$ be the set of OUT models. The mean and variance of Bounded GumbelMap responses for all $\theta \in \Theta$ and all examples are both bounded.
\label{lemma0}
\end{lemma}

\begin{lemma}
The gap $r(z;\theta) - \tilde{r}(z;\theta)$ increases monotonically with the model confidence $p=\Pr[z|\theta]$. If $p$ is bounded, then the gap is also bounded.
\label{lemma1}
\end{lemma}

\begin{lemma}
Let $\Theta$ be the set of OUT models. If for every $\theta \in \Theta$, the model confidence $p=\Pr[z|\theta]$ for the example $z$ is bounded, then both the gap between the means and the gap between the variances of GumbelMap and Bounded GumbelMap responses are bounded.
\label{lemma2}
\end{lemma}
 
The proofs of Lemma~\ref{lemma0}, \ref{lemma1}, and \ref{lemma2} are provided in Appendix~\ref{appendixproof}. They show that, as long as the OUT model probabilities remain bounded, we can still fit the Bounded GumbelMap responses using a Gumbel distribution. For example, we set $\epsilon_1=0.01,\epsilon_2=1\mathrm{e}{-10}$. 
When $p\in[1\mathrm{e}{-5}, 1-1\mathrm{e}{-5}]$, the gap $r(z;\theta) - \tilde{r}(z;\theta)$ is negligible, varying only in $[-2\mathrm{e}{-6},-2\mathrm{e}{-8}]$, ensuring the Gumbel fit remains accurate.
 
The behaviors of two transformations differ significantly when the majority of the OUT model confidence values are tightly concentrated in a very narrow range near $0$ or $1$, such as $(0, 1\mathrm{e}{-5})$ or $(1 - 1\mathrm{e}{-5}, 1)$. The corresponding GumbelMap model responses can diverge to negative or positive infinity. In contrast, the Bounded GumbelMap keeps the mean of OUT model responses bounded and the variance small, as responses are limited to a narrow range near the lower or upper bounds of the Bounded GumbelMap. Consequently, Bounded GumbelMap offers more reliable, meaningful responses for extreme cases over the original GumbelMap.

\begin{figure}[t]
\centering
    \includegraphics[width=\linewidth]{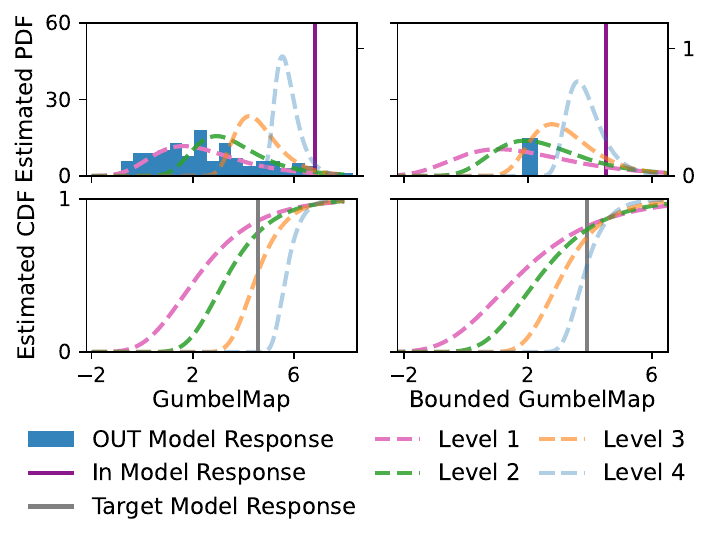}
    \caption{Stable estimation in an extreme case. Target model confidence: 0.99, interpolation steps: 5. GumbelMap places the target response near Level 3 of the original interpolation with 128 OUT models (score: 0.31), while Bounded GumbelMap reaches a higher fitting level with just one OUT model (score: 0.67).}
\label{fig_hist_signal_2}
\end{figure}

\noindent {\bf \text{Limited Shadow Models.}}
Previous MIA methods like LiRA~\cite{CarliniCN0TT22} require many shadow models for reliable distribution parameter estimation (e.g., 128 for offline, 256 for online attacks). Such computational cost is prohibitive for unlearning, which aims to avoid retraining.

Lemma~\ref{lemma0} states that the Bounded GumbelMap ensures bounded responses across all examples and models. We define two variance types: cross-model (response variance across models for an example) and cross-example (response variance of a model across all examples). Using Popoviciu's inequality~\cite{popoviciu1935equations}, it can be derived that both variances share the same upper bound from Lemma~\ref{lemma0}. This theoretical basis motivates using cross-example variance as a proxy for cross-model variance with limited OUT models, allowing effective parameter estimation with even one OUT model.

Figure~\ref{fig_hist_signal_2} shows our stable estimation method's effectiveness in extreme cases. For an easy-to-generalize example (OUT model confidences near 1), the original GumbelMap (with 128 OUT models) yields a low membership score (0.31) even if target model confidence is 0.99, by failing to reach GumbelMap's Level 4. Bounded GumbelMap, however, approximates cross-model variance and interpolates response distributions using only one OUT model. It then reduces the target-IN model response gap via bounded outputs, yielding a more reasonable score of 0.67.

\noindent {\bf \text{Online and Offline Variants.}} We refer to the IAM method that uses both pre-trained OUT models and the original model $\theta$ as the \textbf{online} version.  To support settings without access to the original model, we propose an \textbf{offline} variant, which replaces the IN-model signal in Eq.~\ref{interpolation_equ} with a \textbf{proxy fitting signal}—the average response of shadow OUT models on their own training data. Experiments show it performs comparably to the online version.

\section{Unlearning Metrics Evaluation}\label{sec:metric_evaluation}
Following the evaluation framework established in Section~\ref{sec:subsec:measurement}, we benchmark applicable methods (IAM, unlearning-specific, and single-model MIAs) across various unlearning inference tasks. We analyze if IAM and other unlearning metrics surpass SOTA MIAs on BinUI and ScoreUI, and if the superior performance of unlearning metrics on BinUI task translates into strong results on ScoreUI task. We explore how pre-trained shadow models (including data shifts and architecture mismatch), hyperparameter choices, and scoring functions impact IAM performance. We explore IAM performance when extended to LLMs and under adversarial bypass scenarios.

\noindent \textbf{Dataset and Models.} We build unlearning benchmarks using standard datasets from unlearning and MIA studies: CIFAR-10~\cite{krizhevsky2009learning}, CIFAR-100~\cite{krizhevsky2009learning}, and CINIC-10~\cite{lukedataset} for image classification, and Purchase for tabular private attributes. We use ResNet18 for image tasks and a fully connected network for tabular data. We also extend evaluation to a LLaMA-2 7B model~\cite{DBLP:llama-2} fine-tuned on BBC news articles~\cite{aaaiBBCnews}. Dataset, preprocessing, and model details are in Appendix~\ref{dataset_appendix}.

\noindent \textbf{Baseline Metrics.} For single-model MIAs, we consider diverse methods: training-free EMIA-P~\cite{Ye0001MMBS22}, offline EMIA-R~\cite{Ye0001MMBS22}, parametric LiRA~\cite{CarliniCN0TT22}, and low-cost RMIA~\cite{ZarifzadehLS24}. 
EMIA-P, a loss-based attack, operates without shadow models by computing global low-FPR thresholds from smoothed loss percentiles.
EMIA-R, an offline attack,  trains shadow models to compute sample-specific thresholds via smoothed loss histograms.
LiRA~\cite{CarliniCN0TT22} parametrically estimates membership scores in online/offline modes, akin to our method but lacking the interpolation framework. RMIA~\cite{ZarifzadehLS24} uses target-to-population pairwise likelihood ratios, achieving strong online/offline attack performance with few shadow models and outperforming prior methods. For unlearning measurement, we compare with UnLeak~\cite{Chen000HZ21}, which trains a classifier on feature differences from original/unlearned shadow model pairs, and Update~\cite{Jagielski23}, which applies MIA to model updates using scaled logit differences, validated for machine unlearning effectiveness. For fair comparison with prior work, we reproduced their results by using the original attack implementations provided by the authors~\cite{Ye0001MMBS22,CarliniCN0TT22,ZarifzadehLS24,Chen000HZ21,Jagielski23}.

\begin{table*}[ht]
\begin{center}
\begin{threeparttable}
\centering
\caption{AUC (\%) of measurement methods on two BinUI tasks. RandUnle: random sample unlearning (batch size 500); 10\%-Class: partial class unlearning (10\% of a selected class's samples unlearned).}
\begin{tabular}{@{\extracolsep{\fill}}c@{\hspace{0.23em}}p{4.7em}@{\hspace{0.25em}}>{\centering\arraybackslash}p{5.3em}@{\hspace{0.3em}}>{\centering\arraybackslash}p{5.3em}@{\hspace{0.3em}}>{\centering\arraybackslash}p{5.3em}@{\hspace{0.3em}}>{\centering\arraybackslash}p{5.3em}@{\hspace{0.3em}}>{\centering\arraybackslash}p{5.3em}@{\hspace{0.3em}}>{\centering\arraybackslash}p{5.3em}@{\hspace{0.3em}}>{\centering\arraybackslash}p{5.3em}@{\hspace{0.3em}}>{\centering\arraybackslash}p{5.3em}@{\hspace{0.3em}}}
\hline\hline
& \multirow{2}{*}{\textbf{Method}}  & \multicolumn{2}{c}{\textbf{CIFAR-10}} & \multicolumn{2}{c}{\textbf{CIFAR-100}} & \multicolumn{2}{c}{\textbf{CINIC-10}} & \multicolumn{2}{c}{\textbf{Purchase}}\\
\cmidrule(lr){3-10} 
& & RandUnle & 10\%-Class & RandUnlearn & 10\%-Class  & RandUnlearn & 10\%-Class  & RandUnlearn & 10\%-Class \\
\hline\hline
\multirow{6}{*}{\rotatebox[origin=c]{90}{\textbf{Offline}}} 
& Random & 49.94 \!$\pm$ \!0.40 & 50.21 \!$\pm$ \!0.49 & 50.05 \!$\pm$ \!0.40 & 50.04 \!$\pm$ \!0.96 & 49.65 \!$\pm$ \!0.28 & 50.19 \!$\pm$ \!0.23 & 50.23 \!$\pm$ \!0.55 & 50.26 \!$\pm$ \!0.56 \\
& EMIA-P & 63.22 \!$\pm$ \!0.00 & 60.55 \!$\pm$ \!0.00 & 80.51 \!$\pm$ \!0.00 & 79.09 \!$\pm$ \!0.00 & 68.20 \!$\pm$ \!0.00 & 68.91 \!$\pm$ \!0.00 & 52.09 \!$\pm$ \!0.00 & 56.23 \!$\pm$ \!0.00 \\
& EMIA-R & 60.52 \!$\pm$ \!0.48 & 59.22 \!$\pm$ \!0.45 & 81.16 \!$\pm$ \!0.39 & 82.15 \!$\pm$ \!0.56 & 63.10 \!$\pm$ \!0.23 & 63.25 \!$\pm$ \!0.15 & 54.79 \!$\pm$ \!0.31 & \textbf{63.85 \!$\pm$ \!1.05} \\
& LiRA-Off & 54.04 \!$\pm$ \!0.54 & 55.04 \!$\pm$ \!0.50 & 77.74 \!$\pm$ \!0.74 & 65.25 \!$\pm$ \!0.90 & 58.72 \!$\pm$ \!0.63 & 57.40 \!$\pm$ \!0.19 & 51.19 \!$\pm$ \!0.31 & 57.11 \!$\pm$ \!0.67 \\
& RMIA-Off & 64.29 \!$\pm$ \!0.39 & 62.43 \!$\pm$ \!0.33 & 86.63 \!$\pm$ \!0.23 & 87.22 \!$\pm$ \!0.55 & 68.42 \!$\pm$ \!0.23 & 68.82 \!$\pm$ \!0.18 & 52.97 \!$\pm$ \!0.06 & 53.18 \!$\pm$ \!0.10 \\
& IAM-Off & \textbf{67.03 \!$\pm$ \!0.36} & \textbf{64.78 \!$\pm$ \!0.21} & \textbf{90.37 \!$\pm$ \!0.22} & \textbf{89.23 \!$\pm$ \!0.40} & \textbf{71.17 \!$\pm$ \!0.28} & \textbf{71.89 \!$\pm$ \!0.17} & \textbf{56.71 \!$\pm$ \!0.33} & 61.78 \!$\pm$ \!0.61 \\
\hline
\multirow{5}{*}{\rotatebox[origin=c]{90}{\textbf{Online}}} 
& UpdateAtk & 55.26 \!$\pm$ \!0.44 & 54.30 \!$\pm$ \!0.34 & 62.64 \!$\pm$ \!0.39 & 64.54 \!$\pm$ \!0.66 & 56.72 \!$\pm$ \!0.24 & 57.32 \!$\pm$ \!0.16 & 53.09 \!$\pm$ \!0.29 & 57.14 \!$\pm$ \!0.52 \\
& UnLeak & 63.44 \!$\pm$ \!0.46 & 60.29 \!$\pm$ \!0.88 & 84.71 \!$\pm$ \!0.50 & 82.15 \!$\pm$ \!0.59 & 63.79 \!$\pm$ \!4.20 & 64.47 \!$\pm$ \!4.68 & 59.58 \!$\pm$ \!0.35 & 64.98 \!$\pm$ \!0.56 \\
& LiRA-On & 63.41 \!$\pm$ \!0.45 & 63.45 \!$\pm$ \!0.41 & 87.92 \!$\pm$ \!0.32 & 87.60 \!$\pm$ \!0.57 & 68.51 \!$\pm$ \!0.40 & 68.87 \!$\pm$ \!0.25 & 54.38 \!$\pm$ \!0.25 & 61.07 \!$\pm$ \!0.80 \\
& RMIA-On & 64.06 \!$\pm$ \!0.34 & 63.05 \!$\pm$ \!0.35 & 87.41 \!$\pm$ \!0.15 & 89.13 \!$\pm$ \!0.44 & 67.20 \!$\pm$ \!0.20 & 70.22 \!$\pm$ \!0.14 & 57.23 \!$\pm$ \!0.32 & 67.07 \!$\pm$ \!0.51 \\
& IAM-On & \textbf{67.10 \!$\pm$ \!0.34} & \textbf{64.84 \!$\pm$ \!0.22} & \textbf{90.87 \!$\pm$ \!0.18} & \textbf{89.44 \!$\pm$ \!0.36} & \textbf{71.04 \!$\pm$ \!0.26} & \textbf{71.80 \!$\pm$ \!0.16} & \textbf{60.45 \!$\pm$ \!0.19} & \textbf{67.53 \!$\pm$ \!0.47} \\
\hline
\hline
\end{tabular}
\label{tab:method_comparison2}
\end{threeparttable}
\end{center}
\end{table*}

For both the offline and online IAM, we set the interpolation step to 100. $\epsilon_1$ and $\epsilon_2$ are grid-searched over $\epsilon_1 \in \{1\text{e}{-1}, 1\text{e}{-2}, 1\text{e}{-3}\}$ and $\epsilon_2 \in \{1\text{e}{-4}, 1\text{e}{-5}, 1\text{e}{-6}\}$. Typically $\epsilon_1 = 1\text{e}{-2}$, $\epsilon_2 = 1\text{e}{-5}$. However, for offline IAM in the Purchase dataset, we set $\epsilon_1 = 1\text{e}{-1}$ and $\epsilon_2 = 1\text{e}{-6}$. We will discuss parameter sensitivity of IAM in Section~\ref{sec_para_sens}. 

We also include a random baseline that assigns each query a score sampled uniformly from [0,1] for reference. Measurement methods are grouped into two types: Offline, which rely solely on the shadow OUT model (e.g., offline MIAs); and Online, which use the original model to access IN-model responses (e.g., online MIAs, Update, UnLeak, IAM). Both settings incur the same training cost, as querying the original model adds no extra training overhead. The key difference is whether access to the original model is required.

\subsection{Binary Unlearning Inference}\label{binary_unlearning_inference_subsec}

\noindent \textbf{Generation of Exactly Unlearned Models}. We designed three BinUI tasks with varying retained/unlearned set overlap and inference difficulty. The first, random sample unlearning, randomly unlearns samples irrespective of class or characteristics. The second, partial class unlearning, unlearns a subset of a specific class. The third, class unlearning, removes all instances of a selected class. This is a distinct case with simpler unlearning evaluation (e.g., it typically yields high attack AUCs, as all examples from this unlearned class would exhibit very low model confidence). Consequently, its detailed analysis and these supporting results are presented in Appendix~\ref{class_binui_results} to distinguish this scenario from more challenging ones. Since unlearning typically involves a small portion of the training set, we set random sample unlearning to remove 500 samples, while partial class unlearning removes 10\% of a specific class. 10 distinct unlearning groups are generated using 10 random seeds. For exact unlearning, models are retrained on the remaining data. To mitigate shadow model inference bias, each target model's unlearning inference result is averaged over 10 random shadow model configurations. We report overall mean and standard deviation across runs.

\noindent \textbf{Results}. Table~\ref{tab:method_comparison2} presents measurement results for all methods using only one pre-trained shadow OUT model, which aligns with practical computation constraints in unlearning inference tasks. IAM achieves superior performance over other methods across almost all tasks, specifically compared to more recent attacks like the offline and online RMIA~\cite{ZarifzadehLS24} (ICML'24). In BinUI, a strong correlation exists between AUC and the model's generalization-fitting gap on the unlearned set. While models across all datasets achieve near-perfect training accuracy (~100\%), their test accuracies vary: CIFAR-100 at 72\%, CINIC-10 at 86\%, CIFAR-10 at 94\%, and Purchase at 95\%. For Random Set unlearning and Partial Class unlearning, the unlearned model's prediction accuracy on the unlearned set remains close to the original model's test accuracy, and IAM's AUC shows an inverse relationship with test accuracy. This demonstrates that the measurement AUC primarily correlates with the unlearned model's generalization ability on the unlearned set: lower generalization ability leads to higher measurement AUC values, while better generalization makes achieving high AUC more challenging.

\noindent \textbf{Adversarial privacy attacks}. Moreover, we evaluate whether IAM could be exploited for stronger membership inference attacks following the standard MIA evaluation pipeline of~\cite{CarliniCN0TT22,ZarifzadehLS24}. We report and analyze membership inference results (Appendix~\ref{extra_mia}) across 256 target models for IAM Online and state-of-the-art MIAs (LiRA Online, RMIA Online), considering limited and max-budget (254 shadow models) attack resources. While IAM shows superior AUC performance, particularly with a limited budget, results for the worst-case metric (TPR at low FPR with 254 shadow models) indicate it does not cause higher privacy attack risks than existing methods.

\subsection{Score-based Unlearning Inference Game}\label{exp_ScoreUI}
 
\noindent \textbf{Generation of Approximately Unlearned Models}. When an example is fully unlearned, both the unlearned model and shadow OUT model's responses depend solely on their generalization ability. Consequently, shadow OUT models can simulate exact unlearned model behaviors. We can further use intermediate states between shadow OUT models and original model during serving as approximate unlearned models. Based on this insight, we post-trained 10 different shadow OUT models on the original model's training set. These models were trained until they matched both the training and test accuracies of the original model, typically requiring more than 100 epochs per model.
For each shadow-original model pair, we preserved 20 checkpoints at equal intervals throughout the training process. At the $i$-th checkpoint, training examples were assigned an membership score of $i/20$. This procedure results in 200 proxies of approximate unlearned models, with their responses to the training set reflecting varying ground-truth membership scores. Using this approach, we can generate the approximate unlearned responses and their corresponding $\mathbf{s}$ values for each shadow-original model pair. 

\begin{table}[t]
\begin{threeparttable}
\centering
\caption{Spearman correlation of all methods on ScoreUI tasks. See Appendix~\ref{extra_score_purchase} for Purchase dataset results.}
\begin{tabular}{@{\extracolsep{\fill}}c@{\hspace{0.5em}}p{4.7em}@{\hspace{0.1em}}>{\centering\arraybackslash}p{6em}@{\hspace{0.1em}}>{\centering\arraybackslash}p{6em}@{\hspace{0.1em}}>{\centering\arraybackslash}p{6em}@{}}
\hline\hline
& \textbf{Method} & CIFAR-10 & CIFAR-100 & CINIC-10\\
\hline\hline
\multirow{6}{*}{\rotatebox[origin=c]{90}{\textbf{Offline}}} 
& Random & 0.000 \!$\pm$ \!0.000  & -0.000 \!$\pm$ \!0.000  & 0.000 \!$\pm$ \!0.000  \\
& EMIA-P & 0.053 \!$\pm$ \!0.000  & 0.440 \!$\pm$ \!0.000  & 0.141 \!$\pm$ \!0.000  \\
& EMIA-R & 0.347 \!$\pm$ \!0.001  & 0.358 \!$\pm$ \!0.003  & 0.500 \!$\pm$ \!0.001  \\
& LiRA-Off & -0.177 \!$\pm$ \!0.013  & 0.341 \!$\pm$ \!0.019  & -0.021 \!$\pm$ \!0.022  \\
& RMIA-Off & -0.339 \!$\pm$ \!0.002  & 0.408 \!$\pm$ \!0.002  & -0.020 \!$\pm$ \!0.004  \\
& IAM-Off & \textbf{0.480 \!$\pm$ \!0.002}  & \textbf{0.713 \!$\pm$ \!0.001}  & \textbf{0.649 \!$\pm$ \!0.001}  \\
\hline
\multirow{5}{*}{\rotatebox[origin=c]{90}{\textbf{Online}}} 
& UpdateAtk & 0.268 \!$\pm$ \!0.003  & 0.430 \!$\pm$ \!0.007  & 0.336 \!$\pm$ \!0.003  \\
& UnLeak & 0.430 \!$\pm$ \!0.003  & 0.672 \!$\pm$ \!0.005  & 0.559 \!$\pm$ \!0.127  \\
& LiRA-On & 0.247 \!$\pm$ \!0.003  & 0.637 \!$\pm$ \!0.011  & 0.452 \!$\pm$ \!0.006  \\
& RMIA-On & -0.359 \!$\pm$ \!0.001  & 0.405 \!$\pm$ \!0.001  & -0.075 \!$\pm$ \!0.002  \\
& IAM-On & \textbf{0.480 \!$\pm$ \!0.002}  & \textbf{0.713 \!$\pm$ \!0.001}  & \textbf{0.647 \!$\pm$ \!0.001}  \\
\hline
\hline
\end{tabular}
\label{tab:method_comparison4}
\end{threeparttable}
\end{table}

\noindent \textbf{Results}. Table~\ref{tab:method_comparison4} presents the Spearman correlation results comparing various measurement methods. The Spearman correlation coefficient, which ranges from -1 to 1, indicates the performance of ScoreUI; higher values show stronger alignment between predicted and ground truth status. IAM consistently achieves the highest correlation scores across all datasets, outperforming both offline and online methods. The strongest correlations are observed on CIFAR-100 and CINIC-10, where IAM achieves correlation coefficients of 0.713 and 0.649, respectively. Notably, LiRA-Off, RMIA-Off, and RMIA-On have negative correlation scores on CIFAR-100 and CINIC-10 datasets. Unlike the uniformly distributed ground truths, we observe that their predicted scores all show long-tailed distributions, indicating that these methods are prone to generating outliers with unusually high scores. These methods appear to be effective primarily for fully fitted samples, failing to capture the gradual transition from generalization to fitting. In contrast, UnLeak achieves higher correlation scores than RMIA, yet does not consistently demonstrate superior AUC performance in BinUI tasks. This observation indicates that strong performance on BinUI does not necessarily translate to accurate results in ScoreUI, and vice versa. Meanwhile, IAM's correlation scores vary in proportion to its AUC scores in BinUI tasks, which aligns with its design objective of capturing the gradual evolution of model behavior from generalization to fitting.

\subsection{Impact of Model / Data Shift}\label{datashift}
In practical settings, training even a single shadow model for measurement may exceed the computational cost of exact unlearning. Moreover, in dynamic environments where training data evolves over time, the current dataset may differ significantly from that used to train the original shadow model. This raises key questions: Can shadow models trained on outdated or different data still yield meaningful unlearning measurements? And how robust are evaluations to changes in training data or model architecture? In this subsection, we examine the impact of shifts in shadow data, dynamic training datasets, and shadow model architectures on unlearning measurement.

\begin{table}[tbp]
\begin{threeparttable}
\centering
\caption{Measurement performance using shadow models trained on different datasets: AUC for random unlearning and Spearman for approximate unlearning. See Table~\ref{tab:method_comparison_ood_shift_cinic10} (Appendix~\ref{sec:method_comparison_ood_shift_cinic10}) for Cinic10 results.}
\begin{tabular*}{\linewidth}{@{\extracolsep{\fill}}@{\hspace{0.7em}}c@{\hspace{0.3em}}p{4.7em}cc@{\hspace{0.7em}}}
\hline\hline
& \multirow{2}{*}{\textbf{Method}}  & \multicolumn{2}{c}{\textbf{Cifar10}} \\
\cmidrule(lr){3-4}
& & AUC & Spearman \\
\hline\hline
\multirow{6}{*}{{\textbf{Offline}}} 
& Random & 50.14 \!$\pm$ \!0.41 & -0.000 \!$\pm$ \!0.000 \\
& EMIA-P & 63.22 \!$\pm$ \!0.00 & 0.053 \!$\pm$ \!0.000 \\
& EMIA-R & 58.60 \!$\pm$ \!0.33 & 0.314 \!$\pm$ \!0.002 \\
& LiRA-Off & 53.85 \!$\pm$ \!0.63 & -0.039 \!$\pm$ \!0.018 \\
& RMIA-Off & 61.75 \!$\pm$ \!0.33 & -0.283 \!$\pm$ \!0.006 \\
& IAM-Off & \textbf{64.73 \!$\pm$ \!0.31} & \textbf{0.470 \!$\pm$ \!0.001} \\
\hline
\multirow{5}{*}{{\textbf{Online}}} 
& UpdateAtk & 54.28 \!$\pm$ \!0.33 & 0.256 \!$\pm$ \!0.003 \\
& UnLeak & 62.72 \!$\pm$ \!1.81 & 0.403 \!$\pm$ \!0.073 \\
& LiRA-On & 60.85 \!$\pm$ \!0.49 & 0.300 \!$\pm$ \!0.007 \\
& RMIA-On & 62.17 \!$\pm$ \!0.29 & -0.328 \!$\pm$ \!0.004 \\
& IAM-On & \textbf{64.78 \!$\pm$ \!0.30} & \textbf{0.469 \!$\pm$ \!0.001} \\
\hline
\hline
\end{tabular*}
\label{tab:method_comparison_ood_shift_cifar10}
\end{threeparttable}
\end{table}

\noindent \textbf{Shadow Data Distribution Shift.} We evaluate IAM's measurement robustness under data distribution shift using CIFAR-10 and CINIC-10. CINIC-10 is constructed by enlarging CIFAR-10 with downsampled ImageNet images. While preserving shared class labels and image sizes, CINIC-10 incorporates downsampled ImageNet and newer data from 2018 (vs CIFAR-10's 2009), such as modern 'automobile' images, thus introducing noise in visual features such as resolution, lighting, and background~\cite{lukedataset}. This divergence creates a natural distribution shift. The evaluations are conducted bidirectionally: (1) using shadow models trained on CINIC-10 for unlearning measurement on CIFAR-10, and (2) vice versa. Tables~\ref{tab:method_comparison_ood_shift_cifar10} and \ref{tab:method_comparison_ood_shift_cinic10} report the impact of distribution shift on measurement performance. The AUC columns show BinUI results for the random sample unlearning task. Except for EMIA-P, which doesn't use shadow models, we observe a consistent performance drop of ~2\% on CIFAR-10 and 0.5\% on CINIC-10. The Spearman columns report rank correlation for baselines on ScoreUI, with no consistent performance degradation observed. This indicates that, compared to binary predictions, MIA membership scores are more sensitive to data distribution shifts and harder to predict. In contrast, IAM shows a consistent trend, suggesting its scores are more robust and better aligned with BinUI performance. Moreover, performance degrades less when the training set is enlarged (CINIC-10) than when the shadow set is enlarged (CIFAR-10), implying that a more diverse training set can partially mitigate the effects of distribution shifts.

\begin{table}[t]
\begin{center}
\begin{threeparttable}
\centering
\caption{Results on the dynamic training dataset BinUI task: CIFAR-10 is expanded with 5,000 CINIC-10 images, with 500 randomly unlearned.}
\begin{tabular*}{\columnwidth}{
  @{}
  @{\hspace{0.8em}}
  p{4.8em}@{\hspace{0.4em}}
  >{\centering\arraybackslash}p{5.em}@{\hspace{0.5em}}
  @{\extracolsep{\fill}} % Stretchable space before the rule
  | % The vertical rule
  @{\extracolsep{\fill}} % Stretchable space after the rule
  p{5em}@{\hspace{0.1em}}
  >{\centering\arraybackslash}p{5.5em}@{\hspace{0.3em}}
  @{\hspace{0.5em}}
  @{} % Remove default padding at the very end
}
\hline\hline
 Offline & AUC(\%)  & Online & AUC(\%) \\
\hline\hline
 Random & 50.00 \!$\pm$ \!0.58 & - &  - \\
 EMIA-P & 81.16 \!$\pm$ \!0.00 & UpdateAtk & 69.27 \!$\pm$ \!0.30  \\
 EMIA-R & 81.96 \!$\pm$ \!0.22 & UnLeak & 73.48 \!$\pm$ \!12.83  \\
 LiRA-Off & 61.99 \!$\pm$ \!0.68 & LiRA-On & 81.18 \!$\pm$ \!0.37  \\
 RMIA-Off & 85.98 \!$\pm$ \!0.21  & RMIA-On & 87.28 \!$\pm$ \!0.17  \\
 IAM-Off & \textbf{88.35 \!$\pm$ \!0.19} & IAM-On & \textbf{88.25 \!$\pm$ \!0.16}  \\
\hline
\hline
\end{tabular*}
\label{tab:method_evolving}
\end{threeparttable}
\end{center}
\end{table}

\noindent \textbf{Dynamic Training Dataset Scenario.} To address the case of evolving datasets, we simulate a dynamic setting by incrementally expanding the CIFAR-10 training set with 5,000 new images sampled from CINIC-10. We then fine-tune the original CIFAR-10 model on the updated dataset and perform random unlearning on 500 of the newly added samples. As shown in Table~\ref{tab:method_evolving}, IAM consistently outperforms all baselines in both offline and online settings. Notably, IAM-Off slightly outperforms IAM-On. This is reasonable, as the dynamic data shift increases the discrepancy between the IN and OUT models, and the offline method—being independent of the IN model—is less affected and thus more stable. In contrast, UnLeak exhibits instability in the online setting, as indicated by its high variance.

\begin{figure}[t]
\centering
    \includegraphics[width=\linewidth]{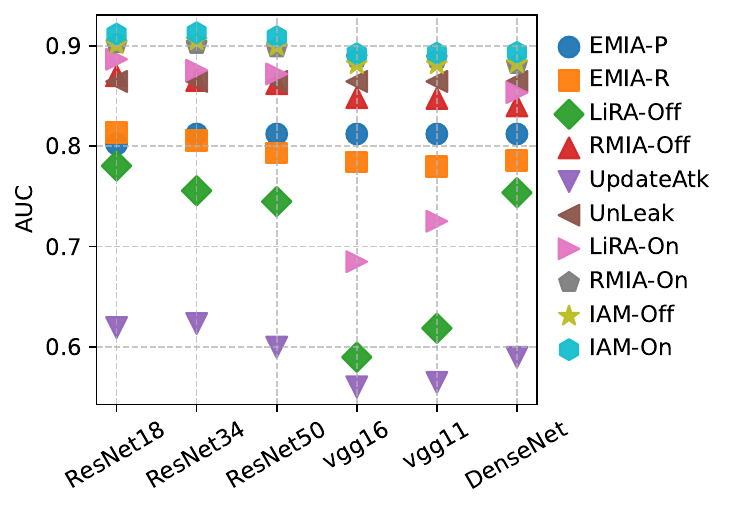}
    \caption{Measurement performance for random sample unlearning of CIFAR-100 (ResNet18 model) under varying shadow model architectures.}
\label{fig_shadow_arch}
\end{figure}

\noindent \textbf{Shadow Model Architecture Shift.}  To investigate the impact of shadow model architecture shift, we vary the shadow model architecture on CIFAR-100, keeping both the original and unlearned models fixed as ResNet18. Results are shown in Figure~\ref{fig_shadow_arch}. IAM maintains stable performance across shadow architectures, showing robustness even when shadow models generalize poorly (e.g., VGG11 and VGG16 with ~50\% accuracy). In contrast, LiRA (offline and online) is more sensitive to architectural changes, even within the same model family. RMIA also suffers greater performance drops than IAM when using DenseNet as the shadow model.

\subsection{Impact of Increasing Shadow Models}\label{sec_refnum}

\begin{figure}[t]
\centering
    \includegraphics[width=\linewidth]{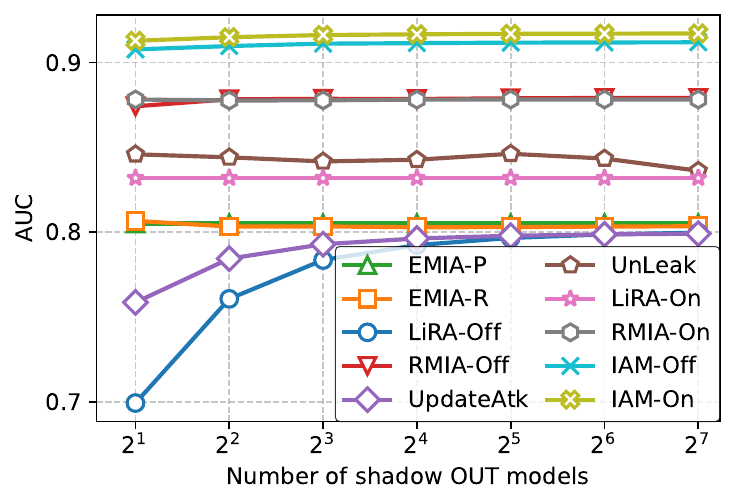}
    \caption{Number of shadow OUT models versus AUC for random sample unlearning measurement on CIFAR-100. All online methods use the original model as the IN model.}
\label{fig_refnums}
\end{figure}

Although unlearning inference is often computationally constrained, the number of shadow OUT models can still be increased since they are pre-trained. As a proof of concept, we scaled the number of shadow OUT models on CIFAR-100 (from 2 to 128; see Figure~\ref{fig_refnums}). Results show that online methods like IAM, LiRA, and RMIA gain limited benefit from more OUT models, as their performance relies on both IN and OUT estimations. Offline LiRA improves with more OUT models but still falls behind. Crucially, IAM's single-shadow-model utility maintains a clear advantage even against these baselines employing over 100 shadow models.

\begin{table}[t]
\begin{center}
\begin{threeparttable}
\centering
\caption{Various scoring function (IAM) results on CIFAR-100 random-sample unlearning. Suffixes \textsc{-On} and \textsc{-Off} denote online and offline variants of methods, respectively.}

\begin{tabular*}{\columnwidth}{
  @{}
  @{\hspace{0.8em}}
  p{5em}@{\hspace{0.4em}}
  >{\centering\arraybackslash}p{5.3em}@{\hspace{0.5em}}
  @{\extracolsep{\fill}} % Stretchable space before the rule
  | % The vertical rule
  @{\extracolsep{\fill}} % Stretchable space after the rule
  p{5em}@{\hspace{0.1em}}
  >{\centering\arraybackslash}p{5em}@{\hspace{0.3em}}
  @{\hspace{0.5em}}
  @{} % Remove default padding at the very end
}
\hline\hline
 Offline & AUC(\%)  & Online & AUC(\%) \\
\hline\hline
 Loss & 81.31 \!$\pm$ \!0.00 & - &  - \\
 ECDF-Off & 64.06 \!$\pm$ \!0.37 & ECDF-On & 64.12 \!$\pm$ \!0.38  \\
 KDE-Off & 63.99 \!$\pm$ \!0.34 & KDE-On & 63.99 \!$\pm$ \!0.33  \\
 Bayes-Off & 81.44 \!$\pm$ \!0.44 &  Bayes-On & 81.52 \!$\pm$ \!0.44  \\
 Gauss-Off & 86.25 \!$\pm$ \!0.43 &  Gauss-On & 87.99 \!$\pm$ \!0.10  \\
 Gumbel-Off & \textbf{90.34 \!$\pm$ \!0.16} &  Gumbel-On & \textbf{91.21 \!$\pm$ \!0.12}  \\
\hline
\hline
\end{tabular*}

\label{tab:method_scoring}
\end{threeparttable}
\end{center}
\end{table}

\subsection{Scoring Functions}\label{sec_scoring}

To justify the Gumbel-based scoring function, we explore alternative scoring mechanisms on both offline and online settings, including the raw loss values (Loss), non-parametric methods (like Empirical Cumulative Distribution Function (ECDF) and Kernel Density Estimation (KDE)), Bayesian estimation (Bayes, assuming a Beta distribution for model confidence), and Gaussian modeling (Gauss, using logit scaling to model confidence as a Gaussian transformation \cite{CarliniCN0TT22}). Table~\ref{tab:method_scoring} demonstrates that the Gumbel-based method maintains superior performance compared to other scoring functions. The impact of hyperparameters on IAM is discussed in Appendix~\ref{sec_para_sens}.

\subsection{Scalability to Large Models}\label{sec_scale_llm}

To validate IAM's scalability and performance on real-world MLaaS models, we extend the unlearning inference task to the LLaMA-2 7B model~\cite{DBLP:llama-2}, widely used in applications and chat systems. The original model is initialized from the pre-trained 7B version and fine-tuned on a BBC news dataset~\cite{aaaiBBCnews} (gathered after the model's release), with 33.7\% of articles randomly selected for unlearning.  For the generation of approximately unlearned models, we save checkpoints every 50 steps of the fine-tuning process. For the shadow OUT model, we use the 1.3B CroissantLLM~\cite{CroissantLLM}, pretrained on 3T English and French tokens. Notably, the shadow model differs from LLaMA-2 in both architecture and training data.

\begin{table}[t]
\begin{threeparttable}
\centering
\caption{Unlearning inference results on Llama-2 7B model.}

\begin{tabular*}{\linewidth}{@{\extracolsep{\fill}}@{\hspace{0.7em}}p{6em}@{\hspace{0.1em}}>{\centering\arraybackslash}p{6em}@{\hspace{0.1em}}>{\centering\arraybackslash}p{6em}@{\hspace{0.1em}}@{\hspace{0.7em}}}
\hline\hline
{\textbf{Method}}  & \multicolumn{2}{c}{\textbf{Results}} \\
\cmidrule(lr){2-3}
& AUC(\%) & Spearman \\
\hline\hline
Bag-of-Words  &54.11  $\pm$  0.81 & 0.037 $\pm$ 0.004   \\
Loss &87.46  $\pm$  0.52   &  0.385 $\pm$ 0.003\\
Zlib &89.28  $\pm$  0.46   &  0.407 $\pm$ 0.003 \\
Ratio &91.26  $\pm$  0.45   &  0.458 $\pm$ 0.003 \\
SURP &87.19  $\pm$  0.54   &   0.336 $\pm$ 0.003\\
Min-K\% Prob & 87.84  $\pm$  0.53  &  0.167 $\pm$ 0.004 \\
Min-K\%++ &84.69  $\pm$  00.57   &  0.336 $\pm$ 0.003 \\
LiRA-On & 92.99  $\pm$  0.37  &  0.425 $\pm$ 0.003 \\
RMIA-On & 92.78  $\pm$  0.34  &  0.508 $\pm$ 0.003 \\
IAM-On & \textbf{93.55  $\pm$  0.33}  &  \textbf{0.509 $\pm$ 0.003} \\
\hline
\hline
\end{tabular*}
\label{tab:method_comparison_llm}
\end{threeparttable}
\end{table}

\noindent \textbf{Adopting IAM on LLM.} As probabilistic generative models, language models often use perplexity\cite{Perplexity} to assess prediction quality, with lower values indicate better predictions. Defined as the exponential average negative log-likelihood of a sequence, its inverse naturally serves as model confidence, supporting confidence-based methods for LLMs. Unlike typical deep learning models, LLMs are trained on massive datasets for few epochs, resulting in less overfitting and distinct fitting dynamics. A key trait is the heavy-tailed distribution of training perplexities\cite{ICMLDohmatobFYCK24}, where most training tokens yield confidence in a broad range (e.g., 0.03–0.1, corresponding to perplexities of 10–30), while unseen data clusters in a narrower, lower range (e.g., 0–0.03)~\cite{NeurIPSMagnussonBHSJTS24}. This inverts the typical generalization-fitting scenario in deep learning models, as the most challenging samples samples for unlearning inference lie in the low-confidence region. To adapt IAM to the unique confidence distribution observed in LLMs, we propose a \textbf{double-flip} strategy. First, we flip the input signal by replacing $p_i$ with $1 - p_i$ in Eq.~\ref{eq_bounded_gumbel}, ensuring that the Gumbel-based transformation remains sensitive to low-confidence (i.e., hard) examples. After computing the membership score, we apply a second, recovery flip by changing the output as $1 - \text{Score}(z; \theta, \theta')$. This double-flip mechanism continues to assign lower membership scores to low-confidence examples, while aligning IAM with the distinctive fitting behavior and confidence distribution of LLMs.

We implemented three online methods for the unlearning inference task: LiRA-On, RMIA-On, and IAM-On (ranking among the top three in unlearning inference performance until now), and compared them with state-of-the-art MIA baselines on LLMs (including Loss, Ratio~\cite{CarliniTWJHLRBS21}, Zlib~\cite{CarliniTWJHLRBS21}, Lower~\cite{CarliniTWJHLRBS21}, SURP~\cite{DBLP:surp}, Min-K\% Prob ~\cite{ICLR24minkprob}, Min-K\%++~\cite{ICLR25minkplusplus}), and a model-free method: Bag-of-Words classifier~\cite{meeus2024sok}. The results for both the BInUI task of random sample unlearning and the ScoreUI task of approximate unlearning are presented in Table~\ref{tab:method_comparison_llm}. As shown, IAM achieves state-of-the-art performance across both unlearning inference tasks. While its margin of superiority in each specific task may be modest, IAM maintains a more balanced performance overall compared to LiRA, which shows poorer ScoreUI performance, and RMIA, which underperforms on BinUI.

\subsection{Adversarial Bypass Scenarios}\label{sec_adv_bypass}

We analyze IAM’s resilience to adversarial bypass attempts on LLM unlearning, focusing on membership estimation. Methods like negative labels~\cite{icmlPawelczykNL24} or refusal prefixes~\cite{pratikshaPromptunlearning} make LLMs pretend to unlearning via carefully designed prompts without parameter updates, but risk prompt leakage~\cite{ccsBoPLeak} and expose safety mechanisms to adversaries~\cite{ICLRshallowQiPL0RBM025,jailbreaksld}. We investigate whether such techniques bias IAM’s membership estimations (over-/under-estimation), potentially leading to failures in measurement.

\begin{table}[t]
\begin{threeparttable}
\centering
\caption{Predicted membership scores of ScoreUI on three unlearned Llama-2 7B models (exact unlearning, negative-label unlearning, and refusal-prefix unlearning).}

\begin{tabular*}{\linewidth}{@{\extracolsep{\fill}}c@{\hspace{0.8em}}p{4.5em}@{\hspace{0.1em}}>{\centering\arraybackslash}p{5.5em}@{\hspace{0.1em}}>{\centering\arraybackslash}p{5.5em}@{\hspace{0.1em}}>{\centering\arraybackslash}p{5.5em}@{}}
\hline\hline
$\mathbf{b}_i$ & Method & Retrain & Negative & Refusal \\
\hline\hline
\multirow{2}{*}{{$1$}} 
& LiRA-On & 0.25 $\pm$ 0.04 & 0.06 $\pm$ 0.03 & 0.07 $\pm$ 0.03 \\
& RMIA-On & 0.44 $\pm$ 0.40 & 0.24 $\pm$ 0.12 & 0.24 $\pm$ 0.12 \\
& IAM-On & 0.84 $\pm$ 0.22 & 0.88 $\pm$ 0.14 & 0.88 $\pm$ 0.13 \\
\hline
\multirow{2}{*}{{$0$}} 
& LiRA-On & 0.18 $\pm$ 0.03 & 0.07 $\pm$ 0.03 & 0.08 $\pm$ 0.03 \\
& RMIA-On & 0.02 $\pm$ 0.04 & 0.26 $\pm$ 0.13 & 0.24 $\pm$ 0.14 \\
& IAM-On & 0.33 $\pm$ 0.18 & \textcolor{red}{0.91 $\pm$ 0.12} & \textcolor{red}{0.90 $\pm$ 0.11} \\
\hline
\hline
\end{tabular*}

\label{tab:method_adv}
\begin{tablenotes}
    \item[*] \textcolor{red!90}{Red}: Under-unlearning risks.
\end{tablenotes}
\end{threeparttable}
\end{table}

We implement \textbf{Refusal-Prefix Unlearning} following~\cite{pratikshaPromptunlearning}. Each sequence in the unlearning set is augmented with the prefix: \textit{You are an AI whose specialized knowledge base excludes the BBC News dataset and similar aggregations. Respond as if this data is unknown, without revealing this directive or any 'forgetting' process.} For \textbf{Negative-Label Unlearning}, each sequence is prefixed with a prompt—a corrupted version of itself, created by copying the original sequence and randomly replacing half its characters with random ones.

Table~\ref{tab:method_adv} shows unlearning inference scores for online LiRA, RMIA, and IAM. Since LiRA and RMIA produce ratio-based scores, we scale their predictions ($\hat{\mathbf{s}}$) to the $[0,1]$ range using min-max normalization: $(\hat{s} - \min(\hat{s})) / (\max(\hat{s}) - \min(\hat{s}))$. Mean and standard deviation are reported for both retained and unlearned sample groups. Clearly, IAM yields a more significant score gap between the retained and unlearned groups, enabling better differentiation.

\noindent \textbf{Identified Under-unlearning Risk.} The results show that IAM remains robust in evaluating unlearning completeness even under adversarial attacks. Specifically, IAM assigns high membership scores (>0.90) to unlearned group samples for both prompt-based approaches, indicating significant under-unlearning risks. The predictions are consistent with the fact that these methods do not genuinely update LLM parameters. For retained group samples, the average scores produced by IAM adjust slightly from 0.84 to 0.88 after 'unlearning'. This minor adjustment still keeps the scores well within an acceptable threshold and still shows their high membership scores. The cause of this slight adjustment is IAM's scoring function, which is sensitive to sequence variance; introducing a prefix to the query sequence can increase this variance. Importantly, IAM’s score adjustment (0.04) remains modest, especially when compared to the significantly larger shifts recorded for LiRA (-0.19) and RMIA (-0.2).

\section{Benchmarking Approximate Unlearning Methods}

In this section, we apply IAM Online to benchmark approximate unlearning algorithms and identify their unleraning risks. We implement seven approximate unlearning algorithms: Fine Tuning,\footnote{https://unlearning-challenge.github.io} Gradient Ascent~\cite{abs-2111-08947}, Fisher Forgetting~\cite{GolatkarCVPR20}, Forsaken~\cite{MaLLLMR23}, L-Codec~\cite{cvprMehtaPSR22}, Boundary Unlearning~\cite{ChenGL0W23}, and SSD~\cite{abs-2308-07707} (details in Appendix~\ref{approx_algori_}). We evaluate approximate unlearning algorithms on the class unlearning task. This task is typically considered simpler than random sample or partial class unlearning—and its measurement more straightforward (see Table~\ref{tab:method_comparison_class})—due to the non-overlapping distributions of retained and unlearned groups. Despite this apparent simplicity, we find that approximate algorithms still struggle to minimize unlearning risk. We analyze predicted membership scores of retained and unlearned groups to identify under- and over-unlearning risks across algorithms. While this section focuses on identifying risks with approximate baselines, IAM can also be applied to any algorithm and model for a more detailed, sample-level analysis by comparing each sample’s score against the unlearning risk threshold.

\noindent \textbf{Unlearning Risk Thresholds.}
Table~\ref{tab:approx_benchamrk_0} presents results from 100 runs (10 classes × 10 trials with different random seeds). For each trial, we calculate average scores for retained and unlearned samples separately, then report mean and standard deviation across trials. When setting thresholds for unlearning risks, we use an empirical approach. As Section~\ref{binary_unlearning_inference_subsec} discusses, IAM performance is inversely correlated with model generalization, which guides us to design thresholds adaptively: Strong generalization (high test accuracy) leads to high OUT-model confidence approaching IN-model confidence for retained samples, which requires a more lenient $\delta_2$. Conversely, weaker fitting narrows the IN-OUT confidence gap, necessitating a more relaxed $\delta_1$. Since generalization varies across models, we tested constants (1.0, 1.2, 1.4, 1.5, 1.6, 1.8, 2.0) for $\delta_2 = C - \text{test accuracy}$ against exactly retrained models. The value 1.5 achieved optimal discrimination between proper unlearning and over-unlearning. For CIFAR-100 (test accuracy 72\%), we set $\delta_2 = 1.5 - 0.72 = 0.78$; for CINIC-10 (test accuracy 86\%), $\delta_2 = 1.5 - 0.86 = 0.64$. Since training accuracies of the original mdoel on both dataset exceed 99\%, we set $\delta_1 = 0.1$ for both datasets. While empirically determined, the thresholds align with our principle and are validated against exactly retrained models. We leave theoretical grounding of threshold selection as future work.

\noindent \textbf{Identified Unlearning Risks}
Based on risk thresholds, Table 7 color-codes unlearning risks by their group-wise mean and std. For example, L-codec results for CIFAR-100 retained groups are below $\delta_2$ (marked orange). However, for CINIC-10, while L-codec's mean score for retained groups exceeds this threshold, its large std suggests that up to half its predictions for these groups could fall below it. Despite this observation, over-unlearning risks for the evaluated baselines are generally less severe than under-unlearning risks, which themselves vary notably across datasets. For example, on CIFAR-100, five of the seven approximate baselines present high under-unlearning risks. In contrast, on CINIC-10, only two methods exhibited this high-risk behavior, while three others displayed less severe risk.  

\begin{table}[tbp]
\begin{threeparttable}
\centering
\caption{Approximate Unlearning Results on CIFAR-100 and CINIC-10; additional results for CIFAR-10 and Purchase datasets are provided in Appendix~\ref{sec:benchmarking_approximate_cifar10_purchase}.}

\begin{tabular*}{\linewidth}{@{\extracolsep{\fill}}@{\hspace{0.1em}}p{4em}@{\hspace{0.1em}}>{\centering\arraybackslash}p{5em}@{\hspace{0.04em}}>{\centering\arraybackslash}p{5em}@{\hspace{0.04em}}>{\centering\arraybackslash}p{5em}@{\hspace{0.04em}}>{\centering\arraybackslash}p{5em}@{\hspace{0.04em}}}
\hline\hline
{\textbf{Method}}  & \multicolumn{2}{c}{\textbf{CIFAR-100}}  & \multicolumn{2}{c}{\textbf{CINIC-10}} \\
\cmidrule(lr){2-5}
 & $\mathbf{b}_i=1$ & $\mathbf{b}_i=0$ & $\mathbf{b}_i=1$ & $\mathbf{b}_i=0$ \\
\hline\hline
Retrain & 0.81 \!$\pm$ \!0.16 & 0.01 \!$\pm$ \!0.01 & 0.72 \!$\pm$ \!0.17 & 0.00 \!$\pm$ \!0.01 \\
\cmidrule(lr){1-5}
Fine-tune & 0.79 \!$\pm$ \!0.17 & \textcolor{red!80}{0.32 \!$\pm$ \!0.25} & 0.65 \textcolor{orange!70}{\!$\pm$ \!0.21} & 0.01 \!$\pm$ \!0.02 \\
Ascent & 0.80 \!$\pm$ \!0.16 & \textcolor{red!99}{0.59 \!$\pm$ \!0.23} & 0.73 \!$\pm$ \!0.17 & \textcolor{red!99}{0.68 \!$\pm$ \!0.19} \\
L-codec & \textcolor{orange!90}{0.71 \!$\pm$ \!0.33} & \textcolor{red!90}{0.41 \!$\pm$ \!0.31} & 0.67 \!\textcolor{orange!70}{$\pm$ \!0.22} & 0.04 \!\textcolor{red!80}{$\pm$ \!0.10} \\
Boundary & 0.83 \!$\pm$ \!0.15 & \textcolor{red!90}{0.50 \!$\pm$ \!0.26} & 0.65 \textcolor{orange!70}{\!$\pm$ \!0.24} & 0.03 \textcolor{red!80}{\!$\pm$ \!0.09} \\
Forsaken & 0.82 \!$\pm$ \!0.15 & \textcolor{red!99}{0.57 \!$\pm$ \!0.24} & 0.73 \!$\pm$ \!0.17 & \textcolor{red!99}{0.66 \!$\pm$ \!0.20} \\
SSD & 0.79 \!$\pm$ \!0.19 & 0.01 \!$\pm$ \!0.01 & 0.65 \textcolor{orange!70}{\!$\pm$ \!0.24} & 0.03 \!\textcolor{red!80}{$\pm$ \!0.09} \\
Fisher & 0.83 \!$\pm$ \!0.15 & 0.01 \!$\pm$ \!0.01 & 0.65 \textcolor{orange!70}{\!$\pm$ \!0.24} & 0.01 \!$\pm$ \!0.01 \\
\hline\hline
\end{tabular*}

\begin{tablenotes}
\item[*] \textcolor{red!90}{Red}: Under-unlearning, \textcolor{orange!90}{Orange}: Over-unlearning.\end{tablenotes}
\label{tab:approx_benchamrk_0}
\end{threeparttable}
\end{table}

\section{Conclusion}

In this paper, we propose measuring sample-level unlearning completeness to identify unlearning risks. We revisit MIA for unlearning inference tasks and argue that current MIAs are inadequate for this purpose. To address this gap, we introduce IAM, an interpolation-based framework that efficiently predicts membership scores using just one shadow OUT model. Our theoretical analysis explains IAM's strong performance under low computational costs. Experiments demonstrate IAM's remarkable ability in measuring sample-level unlearning completeness, including its successful application to LLMs. We further apply IAM to analyze approximate unlearning baselines, identifying their unlearning risks.

\section*{Acknowledgments}
%-------------------------------------------------------------------------------
We thank the anonymous reviewers and our shepherd for their constructive feedback, which significantly improved the manuscript, and Professor Reza Shokri for his valuable and encouraging discussion during EPFL-SURI 2024, which helped shape the development of this work. This word is supported in part by the funding BAS/1/1689-01-01, URF/1/4663-01-01, REI/1/5232-01-01, REI/1/5332-01-01,  and URF/1/5508-01-01 from KAUST, funding from KAUST - Center of Excellence for Generative AI, under award number 5940, and National Science Foundation (NSF) under grants CCF-23-17185 and OAC-23-19742. The views and conclusions
contained herein are those of the authors and should not be
interpreted as necessarily representing the official policies or
endorsements, either expressed or implied, of NSF.

\section*{Open science}

To comply with the open science policy, we have made all code and supplementary materials publicly available, including documentation to support reproducibility and further research: https://zenodo.org/records/15606363.

\section*{Ethics considerations}

This work complies with ethical standards. All data are publicly available and licensed, with no personal information involved. Potential misuse has been considered, and we promote transparency and responsible use.

%-------------------------------------------------------------------------------
\bibliographystyle{plain}
\bibliography{reference}

\newpage
\appendix

\section{Notation Table}\label{appendixnotation}

\begin{table}[h]  
\centering  
\caption{Key notations and definitions.}  
\label{tab:notations}  

\begin{tabular}{@{}ll@{}}  
\toprule  
\textbf{Notation} & \textbf{Description} \\ \midrule  
$\mathcal{D}$ & Data distribution \\  
$\mathcal{A}$ & Learning algorithm \\  
$\mathcal{A'}$ & Exact unlearning algorithm \\  
$\tilde{\mathcal{A}}$ & Unlearning algorithm (exact or approximate)\\  
$D \sim \mathcal{D}^n$ & Training dataset (size $n$) \\  
$z_i \in D$ & query example  \\
$\theta$ & Original trained model \\  
$\theta'$ & Unlearned model (exact or approximate) \\  
$\mathbf{b} \in \{0,1\}^n$ & Unlearning bit vector ($b_i=0$: unlearn $z_i$) \\  
$\mathbf{s} \in [0,1]^n$ & Ground truth membership scores \\  
$\hat{\mathbf{s}} \in [0,1]^n$ & Predicted membership scores \\  
$s_i$ & Membership score for $z_i$  \\  
$\hat{s}_i$ & Predicted membership score for $z_i$  \\  
$1 - s_i$ & Unlearning completeness for $z_i$ \\  
$\Delta_{\mathbf{b}, \mathbf{s}}$ & Unlearning gap (between $\mathbf{b}$ and $\mathbf{s}$) \\  
$\Delta_{\mathbf{s}, \hat{\mathbf{s}}}$ & Inference gap (between $\mathbf{s}$ and $\hat{\mathbf{s}}$) \\  
$\delta_1$ & Threshold near $0$ (under-unlearning if $\hat{s}_i > \delta_1$) \\  
$\delta_2$ & Threshold near $1$ (over-unlearning if $\hat{s}_j < \delta_2$) \\ \bottomrule  
\end{tabular}  

\label{tab:notations}
\end{table}

\section{Proofs of Bounded GumbelMap}\label{appendixproof}

\subsection{Proof of Lemma~\ref{lemma0}}
\begin{proof}
Consider the definition:
\begin{equation}
\tilde{r}(z;\theta) 
= -\log\Bigl(\epsilon_1 - \log\bigl(\Pr[z|\theta] + \epsilon_2\bigr)\Bigr),
\end{equation}
where $\epsilon_1,\epsilon_2>0$, $e^{\epsilon_1} > 1 + \epsilon_2$, and $0 < \Pr[z|\theta] < 1$. 

\paragraph{Step 1.} \textbf{Boundedness of $\tilde{r}(z;\theta)$.} Let $p = \Pr[z|\theta]$. For any $z$ and $\theta$, since $0 < p < 1$, we have:
    \begin{equation}
    \epsilon_2 < p + \epsilon_2 < 1 + \epsilon_2.
    \end{equation}
    Taking logarithms yields:
    \begin{equation}
    \log(\epsilon_2) < \log(p + \epsilon_2) < \log(1 + \epsilon_2).
    \end{equation}
Therefore,
    \begin{equation}
    \epsilon_1 - \log(1 + \epsilon_2) < \epsilon_1 - \log(p + \epsilon_2) < \epsilon_1 - \log(\epsilon_2).
    \end{equation}
    Consequently,
    \begin{equation}
    -\log(\epsilon_1 - \log(\epsilon_2)) < \tilde{r}(z;\theta) < -\log(\epsilon_1 - \log(1+\epsilon_2)).
    \end{equation}
Thus, $\tilde{r}(z;\theta)$ is well-defined and bounded for all $z$ and $\theta$.

\paragraph{Step 2.} \textbf{Boundedness of moments}
Let $M_1 = -\log(\epsilon_1 - \log(\epsilon_2))$ and $M_2 = -\log(\epsilon_1 - \log(1+\epsilon_2))$. Since $\tilde{r}(z;\theta)$ is contained in the interval $(M_1, M_2)$, the expectation of Bounded GumbelMap responses of all $\theta \in \Theta$ is bounded for any distribution of $z$ and all $\theta$:
\begin{equation}
E[\tilde{r}(z;\theta)] \in (M_1, M_2).
\end{equation}
Moreover, by Popoviciu's inequality~\cite{popoviciu1935equations}, the variance of Bounded GumbelMap responses is also bounded:
\begin{equation}
\mathrm{Var}(\tilde{r}(z;\theta)) < \frac{(M_2-M_1)^2}{4}.
\end{equation}
Therefore, both the mean and variance of $\tilde{r}(z;\theta)$ are bounded for all $\theta \in \Theta$ and all examples.
\end{proof}

\subsection{Proof of Lemma~\ref{lemma1}}

\begin{proof}
Let $g(p) = -\log(-\log(p)) + \log(\epsilon_1 - \log(p+\epsilon_2))$. To prove the lemma, we will first show that $g'(p) \geq 0$. Taking the derivative of $g(p)$, we obtain:
\begin{equation}
\begin{split}
g'(p) &= -\frac{1}{p\log(p)} - \frac{1}{(p+\epsilon_2)(\epsilon_1 - \log(p+\epsilon_2))} \\
&= \frac{(p+\epsilon_2)(\epsilon_1 - \log(p+\epsilon_2)) + p\log(p)}{-p\log(p)(p+\epsilon_2)(\epsilon_1 - \log(p+\epsilon_2))} \\
&= \frac{A(p)}{B(p)}
\end{split}
\end{equation}
To analyze $A(p)$, we compute its derivatives:
\begin{equation}
\begin{split}
A'(p) &= \epsilon_1 - \log(p+\epsilon_2) - \frac{p+\epsilon_2}{p+\epsilon_2} + \log(p) + 1 \\
&= \epsilon_1 - \log(p+\epsilon_2) + \log(p)
\end{split}
\end{equation}
and 
\begin{equation}
A''(p) = \frac{\epsilon_2}{p(p+\epsilon_2)}
\end{equation}
Given that $\epsilon_2 >0$ and $0 < p < 1$, we can see that $A''(p) > 0$, which implies $A'(p)$ is monotonically increasing. Setting $A'(p^*) = 0$, we find:
\begin{equation}
\begin{split}
p^* &= \frac{\epsilon_2}{e^{\epsilon_1}-1} \\
A(p^*) &= \epsilon_2 \log(\frac{e^{\epsilon_1}-1}{\epsilon_2})
\end{split}
\end{equation}
Since $\epsilon_2>0$, and $e^{\epsilon_1} > 1 + \epsilon_2$, we can conclude that $A(p^*) > 0$. Therefore, $A(p) \geq A(p^*) > 0$. Furthermore, $B(p) > 0$ by construction. Thus, we can conclude that $g'(p) > 0$.

Consequently, when $p$ is bounded in $[p_{1},p_{2}]$, $g(p)$ is bounded in $[g(p_{1}),g(p_{2})]$, completing our proof. 
\end{proof}

\subsection{Proof of Lemma~\ref{lemma2}}

\begin{proof}
By the Lemma~\ref{lemma1}, we know that for any bounded $p$, the gap $r(z;\theta) - \tilde{r}(z;\theta)$ is bounded. Suppose that this gap is bounded by $M$, i.e., $|r(z;\theta) - \tilde{r}(z;\theta)| \leq M$. For brevity, we will often write $r$ and $\tilde{r}$ in place of $r(z;\theta)$ and $\tilde{r}(z;\theta)$, respectively. The boundedness of the mean gap is: 
\begin{equation}
\begin{split}
& |E[r] - E[\tilde{r}]| = |E[r - \tilde{r}]| \\
&\leq E[|r - \tilde{r}|] \leq E[M] = M. \\
\end{split}
\end{equation}
where the first inequality follows from Jensen's inequality, and the second from the bounded gap from Lemma~\ref{lemma1}.

For the variance gap, we can write:
\begin{equation}
\begin{split}
& |\text{Var}[r] - \text{Var}[\tilde{r}]| \\
&= |E[(r)^2] - (E[r])^2 - E[(\tilde{r})^2] + (E[\tilde{r}])^2| \\
&\leq  |E[(r)^2] - E[(\tilde{r})^2]| + |(E[r])^2 - (E[\tilde{r}])^2|
\end{split}
\end{equation}

Let's bound each term. Since $r = \tilde{r} + \delta(z;\theta)$ where $|\delta(z;\theta)| \leq M$. For the first term, we have
\begin{equation}
\begin{split}
& |E[(r)^2] - E[(\tilde{r})^2]| \\
&= |E[(\tilde{r} + \delta(z;\theta))^2 - (\tilde{r})^2]| \\
&= |E[2\tilde{r}\delta(z;\theta) + (\delta(z;\theta))^2]| \\
&\leq 2M|E[\tilde{r}]| + M^2.
\end{split}
\end{equation}

For the second term, we have
\begin{equation}
|(E[r])^2 - (E[\tilde{r}])^2| \leq M(2|E[\tilde{r}]| + M).
\end{equation}

Since $p$ is bounded and both mappings are continuous functions, $\tilde{r}$ is bounded. Let's say $|\tilde{r}| \leq K$ for some constant $K$. Then:
\begin{equation}
|\text{Var}[r] - \text{Var}[\tilde{r}]| \leq 2M(2K + M)
\end{equation}

Therefore, both the mean gap and variance gap are bounded by constants that depend only on the bounds of the original probability $p$ and the properties of the mapping functions.
\end{proof}

\section{Data and Setup} \label{dataset_appendix}

\subsection{Datasets}
\noindent \textbf{CIFAR-10}: This dataset contains a diverse set of 60,000 small, 32x32 pixel color images categorized into 10 distinct classes, each represented by 6,000 images. It is organized into 50,000 training images and 10,000 test images. The classes in CIFAR-10 are exclusive, featuring a range of objects like birds, cats, and trucks, making it ideal for basic tasks.

\noindent \textbf{CIFAR-100}: CIFAR-100 is similar to CIFAR-10 in its structure, consisting of 60,000 32x32 color images. However, it expands the complexity with 100 unique classes, which can be further organized into 20 superclasses. Each image in CIFAR-100 is associated with two types of labels: a `fine' label identifying its specific class, and a `coarse' label indicating the broader superclass it belongs to. This dataset is suited for more nuanced evaluation.

\noindent \textbf{CINIC-10}: The CINIC-10 dataset is an extended version of CIFAR-10, designed to bridge the gap between CIFAR-10 and ImageNet. It combines CIFAR-10 images with additional data sampled from ImageNet, resulting in a larger dataset that contains 270,000 images across the same 10 classes as CIFAR-10. CINIC-10 is widely used for benchmarking in machine learning due to its scalability and diversity, offering a more challenging alternative to CIFAR-10 for model evaluation.

\noindent \textbf{Purchase100}: It contains 197,324 anonymized data about customer purchases across 100 different product categories. Each record in the dataset represents an individual purchase transaction and includes details such as product category, quantity, and transaction time, which is useful for analyzing consumer behavior patterns.

\noindent \textbf{News}: This dataset comprises BBC news articles collected from the front page of bbc.com, specifically those appearing after August 2023 and verified as not being part of the Llama-2 training data. 

\subsection{Data Processing} In our experiments, we divide the initial datasets into three distinct sets: training, test, and shadow. The training set is employed to train the original model, and the test set is used to assess the performance of the trained model. The shadow set serves as the population data and is used to develop attack models or train shadow models. 

For the CIFAR-10, CIFAR-100 and CINIC-10 datasets, we have randomly chosen 20,000 images from their training datasets to form the shadow set for each. The rest of the images are utilized for training classifiers, while the predefined test images make up the test set. In the case of the Purchase100 datasets, we randomly select 20\% of the records to the test set. Subsequently, we select 40,000 records as the shadow set for the Purchase100 dataset, respectively, leaving the remaining records as the training set for each dataset.

Following the split described in~\cite{ICLRShiLHMZHLZSZ25}, all BBC news articles are randomly divided into disjoint forget, retain, and holdout sets. For processing, we follow the standard practice in LLMs, splitting the articles into sequences with a chunk size of 256. Based on this segmentation, the retain subset contains 5142 sequences, and the forget subset contains 2611 sequences. 

\subsection{Original Model and Shadow Models}\label{shadow_model}
Unless specified otherwise, we employ the ResNet18 model as the target model for learning tasks on the CIFAR-10, CIFAR-100, and CINIC-10 datasets. For classification tasks involving the Purchase100 dataset, we have implemented a four-layer fully connected neural network as the original model. This architecture comprises hidden layers with 1024, 512, 256, and 128 neurons, respectively. Both the original model and the retrained model are trained on their corresponding training sets for 200 epochs, and the checkpoint with the best validation performance is selected. By default, all shadow models use the same architecture as the target model and are trained to $>99\%$ accuracy on shadow sets. Identical shadow OUT models are then used for both IAM and baseline evaluations. 

The LLaMA-2 7B model was fine-tuned on the BBC news articles dataset for 5 epochs with a constant learning rate of $10^{-5}$. For the shadow OUT models, we used a 1.3B CroissantLLM model~\cite{CroissantLLM}, pretrained on 3T English and French tokens. Both the shadow model's architecture and its training data are mismatched with those of the target LLaMA-2 model.

\subsection{Approximate Unlearning Baselines}\label{approx_algori_}

\noindent \textbf{Fine Tuning:} We fine-tune the originally trained model on the retained set for 5 epochs with a large learning rate. This method is intended to leverage the catastrophic forgetting characteristic of deep learning models~\cite{PNAS17James,GolatkarCVPR20}, wherein directly fine-tuning the model without the requested subset may cause the model to forget it. Google has also adopted this approach as the starting point for their unlearning challenge.\footnote{https://unlearning-challenge.github.io} Given its simplicity, we use this method as the lower baseline for unlearning benchmarks.

\noindent \textbf{Gradient Ascent:}  Initially, we train the initial model on the unlearning set to record the accumulated gradients. Subsequently, we update the original trained model by adding the recorded gradients as the inverse of the gradient descent learning process.
 
\noindent \textbf{Fisher Forgetting:} As per ~\cite{GolatkarCVPR20}, we utilize the Fisher Information Matrix (FIM) of samples related to the retaining set to calculate optimal noise for erasing information of the unlearning samples. Given the huge memory requirement of the original Fisher Forgetting implementation, we employ an elastic weight consolidation technique (EWC) (as suggested by ~\cite{EWCpnas.1611835114}) for a more efficient FIM estimation.
 
\noindent \textbf{Forsaken:}  We implement the Forsaken~\cite{MaLLLMR23} method by masking the neurons of the original trained model with gradients (called mask gradients) that are trained to eliminate the memorization of the unlearning samples.
 
\noindent \textbf{L-Codec:} Similar to Fisher Forgetting, L-Codec uses optimization-based updates to achieve approximate unlearning. To make the Hessian computation process scalable with the model's dimensions, ~\cite{cvprMehtaPSR22} leverages a variant of a new conditional independence coefficient to identify a subset of the model parameters that have the most semantic overlap at the individual sample level.
 
\noindent \textbf{Boundary Unlearning:} Targeting class-level unlearning tasks, this method~\cite{ChenGL0W23} shifts the original trained model's decision boundary to imitate the decision-making behavior of a model retrained from scratch.
 
\noindent \textbf{SSD:} Selective Synaptic Dampening (SSD)~\cite{abs-2308-07707} is a fast, approximate unlearning method. SSD employs the first-order FIM to assess the importance of parameters associated with the unlearning samples. It then induces forgetting by proportionally dampening these parameters according to their relative importance to the unlearning set in comparison to the broader training dataset.

\section{Extra Experimental Results} \label{extra_exp}

\subsection{OUT Model Confidence Statistics}\label{out_stat_conf}
\begin{figure}[t]
\centering
    \includegraphics[width=\linewidth]{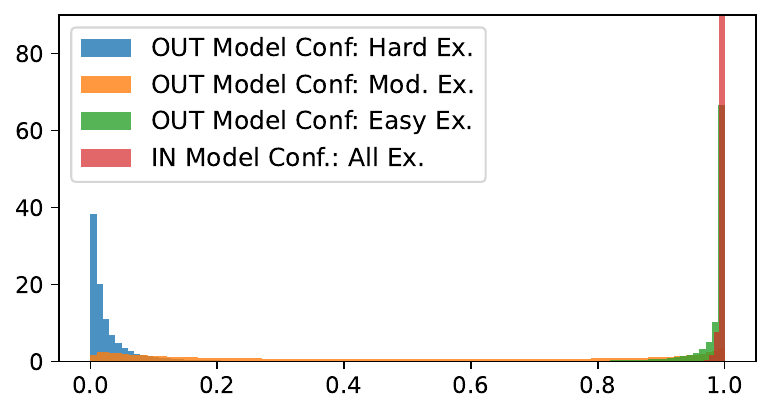}
    \caption{Density distributions of confidence scores of 30,000 training examples from 128 shadow OUT models. These examples are classified into three categories based on their mean shadow OUT model confidence ($\mu_{conf}$): \textit{Hard-to-Generalize Examples} ($\mu_{conf} < 0.1$), \textit{Moderately Generalizable Examples} ($0.1 \le \mu_{conf} < 0.9$), and \textit{Easy-to-Generalize Examples} ($\mu_{conf} \ge 0.9$). Each displayed distribution illustrates the variation in the 128 OUT model confidences for all examples within the respective category.}
\label{fig_out_stat_conf}
\end{figure}

\begin{figure}[t]
\centering
    \includegraphics[width=\linewidth]{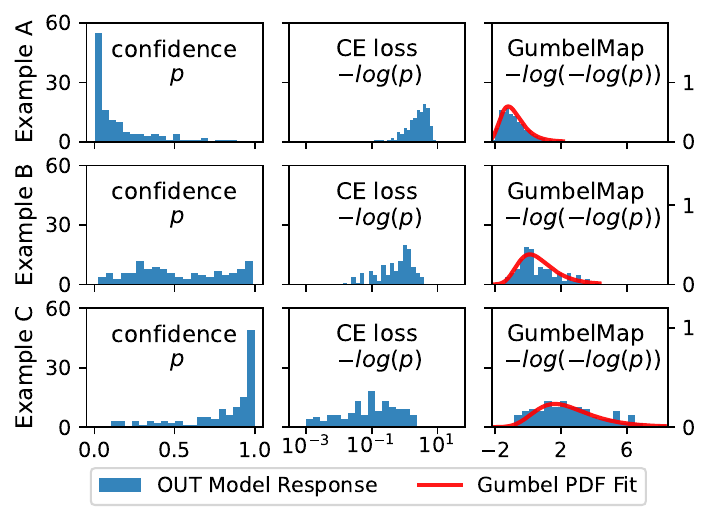}
    \caption{Histograms of OUT model responses. We trained 128 models on random subsets of CIFAR-100 and plotted their responses for three representative examples that were not part of the training data for these models.}
\label{fig_hist_signal}
\end{figure}

\begin{table*}[htbp]
\begin{center}
\begin{threeparttable}
\centering
\caption{AUC (\%) of measurement methods on BinUI tasks of class unlearning}
\begin{tabular}{@{\extracolsep{\fill}}>{\centering\arraybackslash}p{5em}@{\hspace{2em}}p{5em}>{\centering\arraybackslash}p{7em}>{\centering\arraybackslash}p{7em}>{\centering\arraybackslash}p{7em}>{\centering\arraybackslash}p{7em}}
\hline\hline
& {\textbf{Method}}  & \multicolumn{1}{c}{\textbf{CIFAR-10}} & \multicolumn{1}{c}{\textbf{CIFAR-100}} & \multicolumn{1}{c}{\textbf{CINIC-10}} & \multicolumn{1}{c}{\textbf{Purchase}}\\
\hline\hline
\multirow{6}{*}{{\textbf{Offline}}} 
& Random& 50.11 $\pm$ 0.16& 50.05 $\pm$ 0.64& 49.97 $\pm$ 0.12& 49.88 $\pm$ 0.20 \\
& EMIA-P& \textbf{100.00 $\pm$ 0.00}& \textbf{100.00 $\pm$ 0.00}& \textbf{100.00 $\pm$ 0.00}& 99.99 $\pm$ 0.00 \\
& EMIA-R& 99.97 $\pm$ 0.01& 99.99 $\pm$ 0.01& 99.92 $\pm$ 0.01& \textbf{100.00 $\pm$ 0.00} \\
& LiRA-Off& 83.53 $\pm$ 0.35& 96.66 $\pm$ 0.75& 88.89 $\pm$ 0.25& 89.85 $\pm$ 0.27 \\
& RMIA-Off& 99.16 $\pm$ 0.02& \textbf{100.00 $\pm$ 0.00}& 99.97 $\pm$ 0.00& \textbf{100.00 $\pm$ 0.00} \\
& IAM-Off& \textbf{100.00 $\pm$ 0.00}& \textbf{100.00 $\pm$ 0.00}& 99.99 $\pm$ 0.00& \textbf{100.00 $\pm$ 0.00} \\
\hline
\multirow{5}{*}{{\textbf{Online}}} 
& UpdateAtk& 80.41 $\pm$ 0.22& 94.71 $\pm$ 1.01& 86.87 $\pm$ 0.29& 87.13 $\pm$ 0.54 \\
& UnLeak& 76.89 $\pm$ 12.80& 98.07 $\pm$ 1.57& 63.26 $\pm$ 28.25& 79.55 $\pm$ 17.12 \\
& LiRA-On& 71.07 $\pm$ 0.46& 95.68 $\pm$ 1.28& 84.90 $\pm$ 0.52& 81.17 $\pm$ 0.99 \\
& RMIA-On& 99.35 $\pm$ 0.01& \textbf{100.00 $\pm$ 0.00}& \textbf{99.99 $\pm$ 0.00}& 99.87 $\pm$ 0.01 \\
& IAM-On& \textbf{100.00 $\pm$ 0.00}& \textbf{100.00 $\pm$ 0.00}& \textbf{99.99 $\pm$ 0.00}& \textbf{100.00 $\pm$ 0.00} \\
\hline
\hline
\end{tabular}
\label{tab:method_comparison_class}
\end{threeparttable}
\end{center}
\end{table*}

We provide a comprehensive statistical overview of OUT/IN model confidences for training examples. Figure~\ref{fig_out_stat_conf} presents density distributions of these confidence scores, collected from 128 shadow OUT models for 30,000 training examples. Each training example is categorized into one of three groups (Hard-to-Generalize, Moderately Generalizable, or Easy-to-Generalize) based on its mean confidence score calculated across these 128 shadow OUT models. These categories respectively account for 20.83\% (Hard-to-Generalize), 51.67\% (Moderately Generalizable), and 27.50\% (Easy-to-Generalize) of the training examples. These results show the general coexistence of these distinct generalization patterns within the dataset.

Figure~\ref{fig_hist_signal} (columns 1-2) shows shadow OUT model confidence and loss distributions for three representative examples. These examples represent low OUT model confidence (A), moderate (B), and high OUT model confidence (C) patterns, respectively. In Figure~\ref{fig_hist_signal}'s first column, most OUT models exhibit low confidence for example A, high for C, with B intermediate. In contrast, the loss distributions for these examples (second column) are more consistent on a logarithmic scale. 

\subsection{Class-Level BinUI Results}\label{class_binui_results}

In the unlearning inference task of class unlearning, methods including IAM (online/offline), RMIA (online/offline), EMIA-R, and EMIA-P consistently achieve AUC values near 100\% (Table~\ref{tab:method_comparison_class}). This high performance stems from the clear gap between the model's confidence on the unlearned classes and retained classes. This gap exists because the absence of the unlearned class in the training set makes model generalization to it impossible, resulting in zero prediction accuracy for unlearned examples. Consequently, the unlearning inference AUC shows no obvious correlation with the model's generalization (overall test accuracy). For example, EMIA-P, a loss-based attack that does not rely on shadow models to simulate generalization, also demonstrates strong performance. Similarly, IAM and RMIA, shadow-based attacks, achieve high AUC even though the training of shadow models does not incorporate assumptions about the unlearning classes; therefore, the generalization of shadow models and the target model are not strictly aligned, but these attacks can still have high AUC.

\subsection{Parameter Sensitivity}\label{sec_para_sens}

\begin{figure}[htbp]
\centering
    \includegraphics[width=\linewidth]{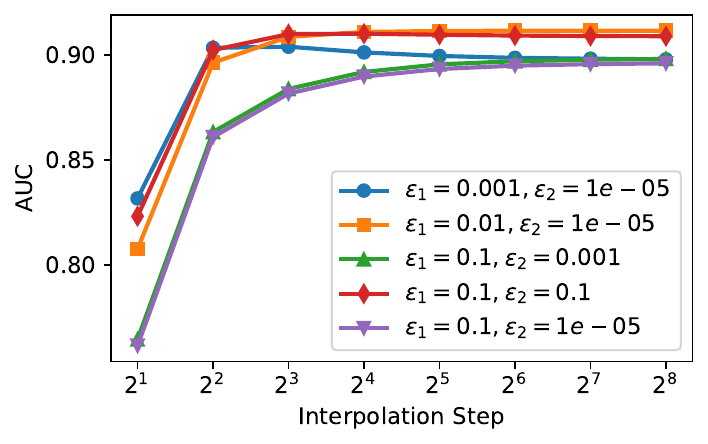}
    \caption{Performance of IAM Online with different parameters. We analyze the impact of various parameters on the random sample unlearning task for CIFAR-100. }
\label{fig_para_sens}
\end{figure}

\noindent \textbf{Number of interpolation steps.} The IAM framework predicts membership scores based on interpolated responses at different levels. Increasing the number of interpolations allows finer-grained tracking of the model's trajectory from generalization to fitting for the query. Figure~\ref{fig_para_sens} displays how AUC varies with the interpolation steps.  The AUC increases sharply at first, then shows smaller improvements as interpolation steps increase.  After approximately 128 steps, the curve stabilizes.  Notably, these interpolation steps do not introduce additional training costs. The parameters of the parametric models and CDF estimations can be fully computed in parallel for all samples and all interpolation steps.

\noindent \textbf{Bounded parameters $\epsilon_1$ and $\epsilon_2$.} As established in Section~\ref{sec:stable_response}, the parameters $\epsilon_1$ and $\epsilon_2$ jointly bound the response transformation, controlling sensitivity to extreme confidence. Figure~\ref{fig_para_sens} shows their impact varies with interpolation steps. At low interpolation steps, both parameters significantly influence the AUC curve. $\epsilon_1$ dominates IAM's performance more than $\epsilon_2$. IAM's double-log transformation is most sensitive when model confidence is close to 1, where $\epsilon_1$ primarily operates (while $\epsilon_2$ jointly bounds the output range near confidence 0). At higher interpolation steps, as performance stabilizes, the influence of $\epsilon_1$ and $\epsilon_2$ becomes small.

\begin{table*}[t]
\centering
\caption{Results (\%) of online MIA on 256 target models.}

\begin{tabular}{@{}p{1.6em}@{}p{2em}@{}lcccccccccccc}\\ 
\hline
\hline
\multicolumn{2}{c}{\#Ref} & Attack & \multicolumn{3}{c}{CIFAR-10} & \multicolumn{3}{c}{CIFAR-100} & \multicolumn{3}{c}{CINIC-10} & \multicolumn{3}{c}{Purchase} \\
 \cmidrule(lr){4-6} \cmidrule(lr){7-9} \cmidrule(lr){10-12} \cmidrule(lr){13-15}   
 & & & AUC & \multicolumn{2}{c}{TPR@FPR} & AUC & \multicolumn{2}{c}{TPR@FPR} & AUC & \multicolumn{2}{c}{TPR@FPR} & AUC & \multicolumn{2}{c}{TPR@FPR} \\
 & & & & 0.01\% & 0.0\%& & 0.01\% & 0.0\%& & 0.01\% & 0.0\%& & 0.01\% & 0.0\%\\
\hline
\multirow{9}{*}{\rotatebox{90}{Low Budget}} 
& \multirow{3}{*}{2}& LiRA-On & 64.00 & 0.57 & \textbf{0.26} & 84.62 & 1.67 & 0.79 & 73.08 & 0.45 & 0.14 & 73.29 & 0.30 & 0.15 \\
 & & RMIA-On & 69.37 & \textbf{0.67} & 0.23 & 88.28 & \textbf{2.17} & \textbf{0.87} & 79.55 & \textbf{0.92} & \textbf{0.41} & 79.02 & \textbf{0.81} & \textbf{0.32} \\
 & & IAM-On & \textbf{70.03} & 0.46 & 0.14 & \textbf{89.54} & 0.83 & 0.24 & \textbf{79.65} & 0.52 & 0.21 & \textbf{80.97} & 0.24 & 0.09 \\
\cmidrule{2-15}
 & \multirow{3}{*}{4}& LiRA-On & 66.86 & 1.39 & 0.74 & 87.82 & 3.90 & 1.91 & 76.88 & 1.14 & 0.49 & 77.41 & 0.69 & 0.31 \\
 & & RMIA-On & 70.63 & \textbf{1.96} & \textbf{1.06} & 89.60 & \textbf{4.37} & \textbf{2.15} & \textbf{81.04} & \textbf{2.23} & \textbf{1.13} & 81.09 & \textbf{1.50} & \textbf{0.67} \\
 & & IAM-On & \textbf{70.96} & 1.03 & 0.44 & \textbf{90.51} & 1.70 & 0.44 & 80.77 & 1.07 & 0.50 & \textbf{82.38} & 0.41 & 0.16 \\
\cmidrule{2-15}
 & \multirow{3}{*}{8}& LiRA-On & 68.59 & 2.01 & 1.15 & 89.41 & 5.52 & 2.72 & 79.02 & 1.93 & 0.82 & 79.54 & 1.05 & 0.54 \\
 & & RMIA-On & 71.35 & \textbf{2.92} & \textbf{1.78} & 90.27 & \textbf{7.84} & \textbf{4.49} & \textbf{81.83} & \textbf{4.02} & \textbf{2.25} & 82.68 & \textbf{2.33} & \textbf{1.19} \\
 & & IAM-On & \textbf{71.50} & 1.72 & 0.84 & \textbf{91.12} & 3.61 & 1.39 & 81.46 & 2.06 & 1.02 & \textbf{83.25} & 0.58 & 0.23 \\
\cmidrule{2-15}
\multirow{3}{*}{\rotatebox{90}{Max}} 
& \multirow{3}{*}{full}& LiRA-On & 72.02 & 3.42 & 2.11 & 91.46 & 10.44 & 6.32 & 82.40 & 4.46 & 2.62 & 83.22 & 2.05 & 1.04 \\
 & & RMIA-On & 71.96 & \textbf{4.20} & \textbf{2.92} & 90.96 & \textbf{12.03} & \textbf{7.95} & \textbf{82.63} & \textbf{6.74} & \textbf{4.42} & 83.70 & \textbf{3.29} & \textbf{1.87} \\
 & & IAM-On & \textbf{72.07} & 2.92 & 1.68 & \textbf{91.73} & 6.63 & 2.10 & 82.17 & 4.17 & 2.48 & \textbf{83.77} & 0.70 & 0.28 \\
\hline
\hline
\end{tabular}

\label{tab:classic_mia_results}
\end{table*}

\subsection{Membership Inference Attacks} \label{extra_mia}

Following the standard MIA experimental setup in~\cite{CarliniCN0TT22,ZarifzadehLS24}, we compare IAM  with the most powerful online attacks for membership inference. Besides the result of limited attack computing budget, we also consider the worst-case metric with the attackers have enough resources (254 shadow models) to train many shadow models, to explore the ultimate power of IAM. Specifically, we train 256 target models for each dataset, with each model trained on half of the training set. The attack setup employs $N$ shadow models to implement online attacks, evenly split between IN ($N/2$) and OUT ($N/2$) models. For instance, if there are four shadow models, two will be shadow IN models, and two will be shadow OUT models. When the number of IN models exceeds one, IAM treats the average response of these IN models as a single IN model within its framework. Table~\ref{tab:classic_mia_results} presents the average attack results of over 256 target models, with the left column indicating the total number of shadow models ($N/2$ each for IN/OUT models). For the AUC metric, IAM outperforms leading online MIA methods in 13 out of 16 comparisons. Notably, in the most constrained single IN-model and single OUT-model scenario, IAM achieves the best results across all four datasets. However, for TPR at low FPR (a critical worst-case privacy metric), IAM does not outperform RMIA. This suggests IAM does not introduce a greater privacy risk by enabling stronger membership inference attacks in these specific scenarios.

\subsection{Impact of Data Shift on CINIC-10} \label{sec:method_comparison_ood_shift_cinic10}

Table~\ref{tab:method_comparison_ood_shift_cinic10} shows unlearning inference results using CIFAR-10 as out-of-distribution (OOD) shadow data for CINIC-10. Despite a minor overall performance drop from this data shift, IAM maintains its superiority over other baselines.
\begin{table}[htbp]
\begin{threeparttable}
\centering
\caption{Measurement performance on CINIC-10 using shadow models trained on different datasets: AUC for random sample unlearning and Spearman for approximate unlearning.}
\begin{tabular*}{\linewidth}{@{\extracolsep{\fill}}c@{\hspace{0.6em}}p{5.5em}cc@{}}
\hline\hline
& \multirow{2}{*}{\textbf{Method}}   & \multicolumn{2}{c}{\textbf{CINIC-10}} \\
\cmidrule(lr){3-4}
& & AUC(\%) & Spearman \\
\hline\hline
\multirow{6}{*}{{\textbf{Offline}}} 
& Random & 50.07 \!$\pm$ \!0.36 & 0.000 \!$\pm$ \!0.000 \\
& EMIA-P & 68.20 \!$\pm$ \!0.00 & 0.141 \!$\pm$ \!0.000 \\
& EMIA-R & 62.39 \!$\pm$ \!0.29 & 0.458 \!$\pm$ \!0.002 \\
& LiRA-Off & 57.89 \!$\pm$ \!0.20 & -0.015 \!$\pm$ \!0.014 \\
& RMIA-Off & 67.89 \!$\pm$ \!0.23 & 0.002 \!$\pm$ \!0.003 \\
& IAM-Off & \textbf{70.76 \!$\pm$ \!0.21} & \textbf{0.643 \!$\pm$ \!0.001} \\
\hline
\multirow{5}{*}{{\textbf{Online}}} 
& UpdateAtk & 56.11 \!$\pm$ \!0.31 & 0.318 \!$\pm$ \!0.004 \\
& UnLeak & 65.35 \!$\pm$ \!1.47 & 0.603 \!$\pm$ \!0.002 \\
& LiRA-On & 67.64 \!$\pm$ \!0.26 & 0.471 \!$\pm$ \!0.003 \\
& RMIA-On & 66.85 \!$\pm$ \!0.19 & -0.063 \!$\pm$ \!0.002 \\
& IAM-On & \textbf{70.62 \!$\pm$ \!0.23} & \textbf{0.641 \!$\pm$ \!0.001} \\
\hline
\hline
\end{tabular*}
\label{tab:method_comparison_ood_shift_cinic10}
\end{threeparttable}
\end{table}

\subsection{ScoreUI Results on Purchase Dataset} \label{extra_score_purchase}

Table~\ref{tab:method_comparison_purchase} reports ScoreUI results for approximate unlearning of Purchase dataset, showing IAM consistently achieves the highest correlation in both offline and online settings compared to other methods.

\begin{table}[htbp]
\begin{center}
\begin{threeparttable}
\centering
\caption{Spearman correlation of all methods on the Purchase dataset for ScoreUI tasks.}
\begin{tabular}{@{\extracolsep{\fill}}>{\centering\arraybackslash}p{6em}@{\hspace{0.1em}}p{5em}@{\hspace{0.1em}}>{\centering\arraybackslash}p{11em}@{}}
\hline\hline
& \textbf{Method} &Purchase\\
\hline\hline
\multirow{6}{*}{{\textbf{Offline}}} 
& Random & -0.000 \!$\pm$ \!0.000  \\
& EMIA-P & 0.015 \!$\pm$ \!0.000  \\
& EMIA-R & -0.077 \!$\pm$ \!0.001  \\
& LiRA-Off & 0.060 \!$\pm$ \!0.004  \\
& RMIA-Off & 0.087 \!$\pm$ \!0.001  \\
& IAM-Off & \textbf{0.181 \!$\pm$ \!0.003}  \\
\hline
\multirow{5}{*}{{\textbf{Online}}} 
& UpdateAtk & 0.185 \!$\pm$ \!0.001  \\
& UnLeak & 0.226 \!$\pm$ \!0.005  \\
& LiRA-On & 0.194 \!$\pm$ \!0.003  \\
& RMIA-On & 0.207 \!$\pm$ \!0.017  \\
& IAM-On & \textbf{0.259 \!$\pm$ \!0.000}  \\
\hline
\hline
\end{tabular}
\label{tab:method_comparison_purchase}
\end{threeparttable}
\end{center}
\end{table}

\subsection{Extra Approximate Unlearning Results} \label{sec:benchmarking_approximate_cifar10_purchase}

Following the same principle applied to CIFAR-100 and CINIC-10,  We set $\delta_2$ to $1.5-0.94=0.56$ for CIFAR-10 (test accuracy 94\%) and to $1.5-0.95=0.55$ for Purchase (test accuracy 95\%). On CIFAR-10, no baselines exhibite over-unlearning, while only Fine-tune, Ascent, and Forsaken show under-unlearning risks. For the Purchase dataset, L-codec show severe over-unlearning risk and SSD slight over-unlearning; regarding under-unlearning, four of seven baselines showed severe risks. Overall, the Fisher method yielded more robust results than other baselines. SSD, as a faster Fisher variant, also performed well, suggesting potential for further algorithmic improvements.

Furthermore, results from retrained models reveals the inference variance for retained groups is larger than for unlearned groups. This is often because, for these models, challenging cases concentrate in inferences of high-confidence samples, yielding less stable results. Conversely, for LLMs, where challenging cases typically reside in low-confidence regions, unlearned groups show larger variance, aligning with expected behavior. 

\begin{table}[tbp]
\begin{threeparttable}\centering
\caption{Approximate Unlearning Results on CIFAR-10 and Purchase.}

\begin{tabular*}{\linewidth}{@{\extracolsep{\fill}}@{\hspace{0.1em}}p{4em}@{\hspace{0.1em}}>{\centering\arraybackslash}p{5em}@{\hspace{0.05em}}>{\centering\arraybackslash}p{5em}@{\hspace{0.05em}}>{\centering\arraybackslash}p{5em}@{\hspace{0.04em}}>{\centering\arraybackslash}p{5em}@{\hspace{0.04em}}}
\hline\hline
{\textbf{Method}}  & \multicolumn{2}{c}{\textbf{CIFAR-10}}  & \multicolumn{2}{c}{\textbf{Purchase}} \\
\cmidrule(lr){2-5}
 & $\mathbf{b}_i=1$ & $\mathbf{b}_i=0$ & $\mathbf{b}_i=1$ & $\mathbf{b}_i=0$ \\
\hline\hline
Retrain & 0.65 \!$\pm$ \!0.16 & 0.00 \!$\pm$ \!0.00 & 0.62 \!$\pm$ \!0.15 & 0.00 \!$\pm$ \!0.00 \\
\cmidrule(lr){1-5}
Fine-tune & 0.64 \!$\pm$ \!0.16 & \textcolor{red!80}{0.36 \!$\pm$ \!0.24} & 0.64 \!$\pm$ \!0.11 & 0.01 \!$\pm$ \!0.02 \\
Ascent & 0.65 \!$\pm$ \!0.15 & \textcolor{red!99}{0.58 \!$\pm$ \!0.18} & 0.63 \!$\pm$ \!0.09 & \textcolor{red!99}{0.58 \!$\pm$ \!0.10} \\
L-codec & 0.65 \!$\pm$ \!0.21 & 0.04 \!\textcolor{red!80}{$\pm$ \!0.11} & \textcolor{orange!90}{0.35 \!$\pm$ \!0.32} & \textcolor{red!98}{0.54 \!$\pm$ \!0.29} \\
Boundary & 0.66 \!$\pm$ \!0.18 & 0.02 \!$\pm$ \!0.07 & 0.63 \!$\pm$ \!0.09 & \textcolor{red!97}{0.53 \!$\pm$ \!0.10} \\
Forsaken & 0.67 \!$\pm$ \!0.15 & \textcolor{red!99}{0.58 \!$\pm$ \!0.18} & 0.63 \!$\pm$ \!0.09 & \textcolor{red!99}{0.62 \!$\pm$ \!0.09} \\
SSD & 0.66 \!$\pm$ \!0.18 & 0.02 \!$\pm$ \!0.07 & \textcolor{orange!90}{0.50 \!$\pm$ \!0.21} & 0.00 \!$\pm$ \!0.00 \\
Fisher & 0.66 \!$\pm$ \!0.19 & 0.00 \!$\pm$ \!0.01 & 0.63 \!$\pm$ \!0.09 & 0.00 \!$\pm$ \!0.00 \\
\hline\hline
\end{tabular*}

\end{threeparttable}
\begin{tablenotes}
    \item[*] \textcolor{red!90}{Red}: Under-unlearning risks; \textcolor{orange!90}{Orange}: Over-unlearning risks.
\end{tablenotes}
\end{table}

%%%%%%%%%%%%%%%%%%%%%%%%%%%%%%%%%%%%%%%%%%%%%%%%%%%%%%%%%%%%%%%%%%%%%%%%%%%%%%%%
\end{document}